\documentclass{article}

\usepackage{PRIMEarxiv}

\usepackage[utf8]{inputenc} % allow utf-8 input
\usepackage[T1]{fontenc}    % use 8-bit T1 fonts
\usepackage{hyperref}       % hyperlinks
\usepackage{url}            % simple URL typesetting
\usepackage{booktabs}       % professional-quality tables
\usepackage{amsfonts}       % blackboard math symbols
\usepackage{nicefrac}       % compact symbols for 1/2, etc.
\usepackage{microtype}      % microtypography
\usepackage{lipsum}
\usepackage{fancyhdr}       % header
\usepackage{graphicx}       % graphics
\graphicspath{{media/}}     % organize your images and other figures under media/ folder

\usepackage{wrapfig}
\usepackage{amsmath}
\usepackage{graphicx}
\usepackage{color}
\usepackage{csvsimple}
\usepackage{datatool}
\usepackage{graphicx}
\usepackage{tikz}
\usepackage{multirow}
\usepackage{amssymb}% http://ctan.org/pkg/amssymb
\usepackage{pifont}% http://ctan.org/pkg/pifont
\usepackage{comment}
\newcommand{\xmark}{--}

\usepackage{array}
\newcolumntype{L}[1]{>{\raggedright\let\newline\\\arraybackslash\hspace{0pt}}m{#1}}
\newcolumntype{R}[1]{>{\raggedleft\let\newline\\\arraybackslash\hspace{0pt}}m{#1}}
\usepackage{algorithm}
\usepackage{algpseudocode}

\usepackage{booktabs}
\usepackage{xspace}
\usepackage{doi}
\makeatletter
\DeclareRobustCommand\onedot{\futurelet\@let@token\@onedot}
\def\@onedot{\ifx\@let@token.\else.\null\fi\xspace}
\def\eg{\emph{e.g}\onedot} \def\Eg{\emph{E.g}\onedot}
\def\etal{\emph{et al}\onedot}

\makeatletter
\def\@onedot{\ifx\@let@token.\else.\null\fi\xspace}
\def\onedot{\@onedot}
  \def\eg{\emph{e.g}\onedot}   \def\Eg{\emph{E.g}\onedot}

  \def\etal{\emph{et al}\onedot}
    
\makeatother

\usepackage{xcolor,pifont}
\newcommand*\colourcheck[1]{%
  \expandafter\newcommand\csname #1check\endcsname{\textcolor{#1}{\ding{52}}}%
}
\newcommand*\colourcross[1]{%
  \expandafter\newcommand\csname #1cross\endcsname{\textcolor{#1}{\ding{55}}}%
}
\colourcheck{green}
\colourcheck{yellow}
\colourcross{yellow}
\colourcross{red}

\title{synth-dacl: Does Synthetic Defect Data Enhance Segmentation Accuracy and Robustness for Real-World Bridge Inspections?
}

\begin{document}
\makeatletter
  \vbox{%
    \hsize\textwidth
    \linewidth\hsize
    \vskip 0.1in
    \@toptitlebar
    \centering
    {\LARGE\sc \@title\par}
    \@bottomtitlebar
    \vskip 0.2in
    \centering

    {\bfseries
    Johannes Flotzinger\textsuperscript{1}, 
    Fabian Deuser\textsuperscript{1}, 
    Achref Jaziri\textsuperscript{2}, 
    Heiko Neumann\textsuperscript{3},\\
    Norbert Oswald\textsuperscript{1}, 
    Visvanathan Ramesh\textsuperscript{2}, 
    Thomas Braml\textsuperscript{1}
    \par}

    \vskip 1.5ex
    \normalsize\rm
    \begin{tabular}{c}
      \textsuperscript{1}University of the Bundeswehr Munich\\
      \textsuperscript{2}Goethe University Frankfurt am Main\\
      \textsuperscript{3}University of Ulm
    \end{tabular}

    \vskip 0.4in \@minus 0.1in
    \center{}
    \vskip 0.2in
  }

\makeatother

\begin{abstract}
Adequate bridge inspection is increasingly challenging in many countries due to  growing ailing stocks, compounded with a lack of staff and financial resources. 
Automating the key task of visual bridge inspection 
-- classification of defects and building components on pixel level --
improves efficiency, increases accuracy and enhances safety in the inspection process and resulting building assessment. 
Models overtaking this task must cope with an assortment of real-world conditions. They must be robust to variations in image quality, as well as background texture, as defects often appear on surfaces of diverse texture and degree of weathering.
dacl10k~\cite{dacl10k} is the largest and most diverse dataset for real-world concrete bridge inspections. However, the dataset exhibits class imbalance, which leads to notably poor model performance particularly when segmenting fine-grained classes such as cracks and cavities.  
This work introduces ``\textit{synth-dacl}'', a compilation of three novel dataset extensions based on synthetic concrete textures. These extensions are designed to balance class distribution in dacl10k and enhance model performance, especially for crack and cavity segmentation.
When incorporating the \textit{synth-dacl} extensions, we observe substantial improvements in model robustness across 15 perturbed test sets. Notably, on the perturbed test set, a model trained on dacl10k combined with all synthetic extensions achieves a 2\% increase in mean IoU, F1 score, Recall, and Precision compared to the same model trained solely on dacl10k.
\end{abstract}

% keywords can be removed
\keywords{Automated Damage Recognition \and Bridge Inspection \and Synthetic Data \and Semantic Segmentation}

%%%%%%%% INTRODUCTION
\section{Introduction}
\label{sec:intro}

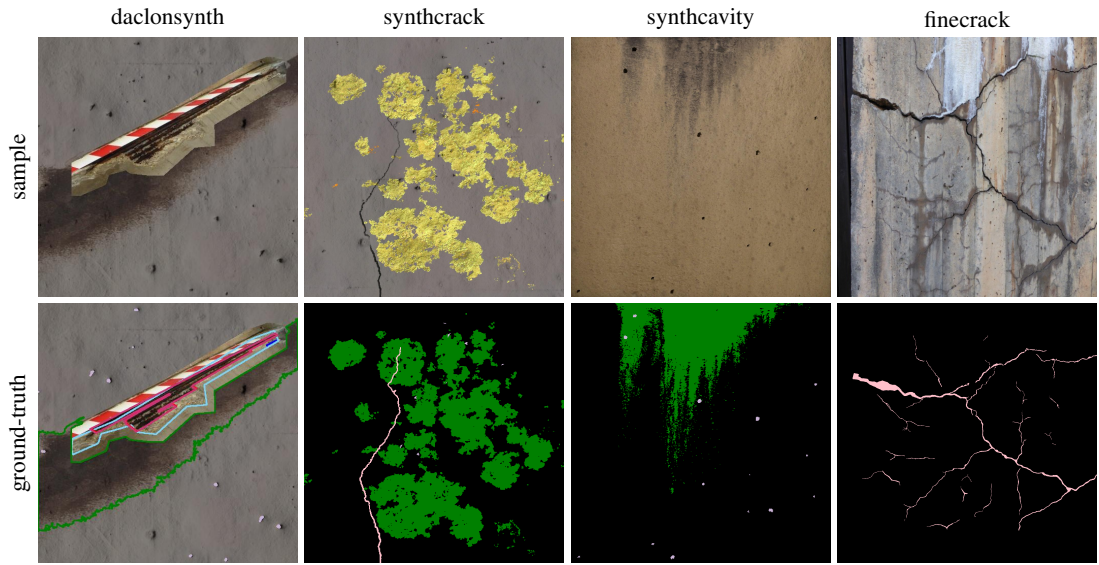
\begin{figure*}[!ht]
\centering
\resizebox{.9\linewidth}{!}{
\begin{tikzpicture}
% Set image width and spacing
\def\imgwidth{3.5cm}  % Image width
\def\xstart{2}  % Starting x position
\def\ystart{8}  % Starting y position
\def\xoffset{3.6}  % Horizontal offset between images
\def\yoffset{-3.6}  % Vertical offset between rows
% Adjusted y-offset for labels
\def\labeloffset{2}  % Additional y-offset for labels to prevent overlap

% First row (Sample)
\node at (\xstart-2, \ystart) {\rotatebox{90}{\small sample}};
\node at (\xstart, \ystart) {\includegraphics[width=\imgwidth]{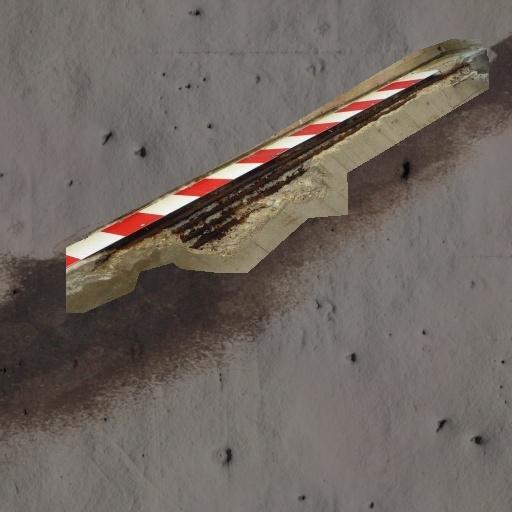}};
\node at (\xstart+\xoffset, \ystart) {\includegraphics[width=\imgwidth]{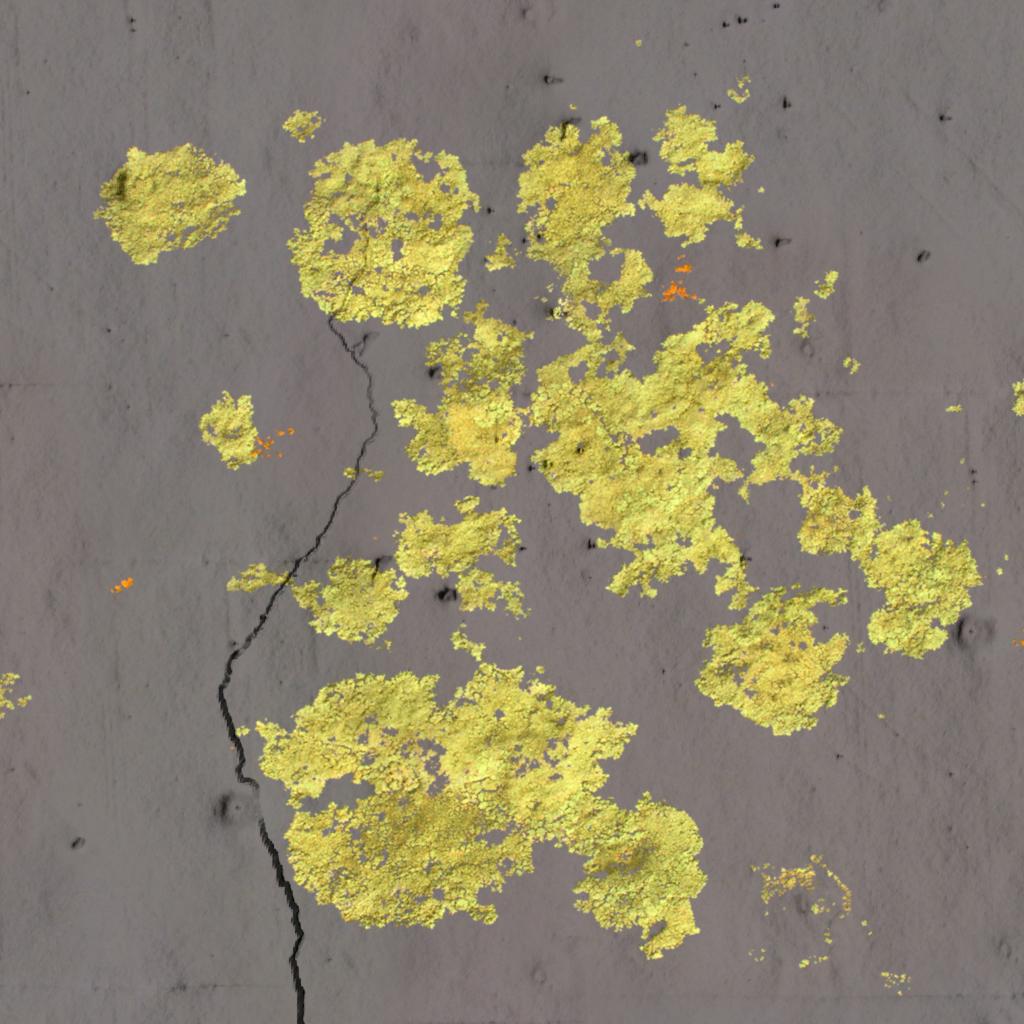}};
\node at (\xstart+\xoffset*2, \ystart) {\includegraphics[width=\imgwidth]{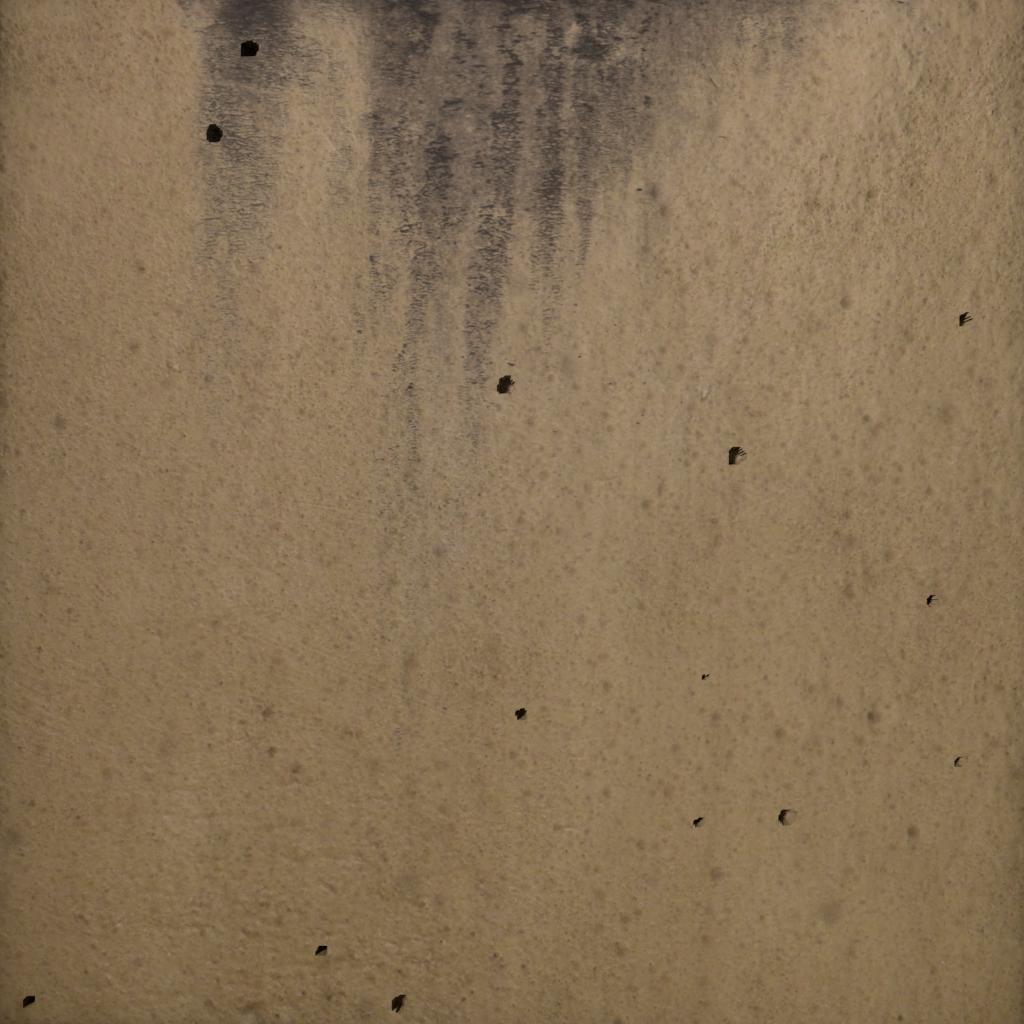}};
\node at (\xstart+\xoffset*3, \ystart) {\includegraphics[width=\imgwidth]{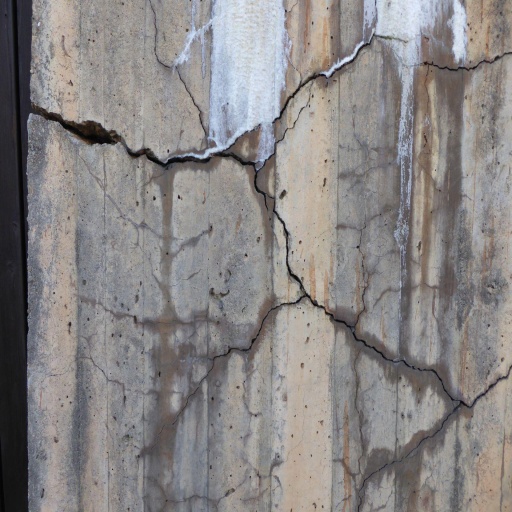}}; % new finecrack sample

% Second row (ground-truth)
\node at (\xstart-2, \ystart+\yoffset) {\rotatebox{90}{\small ground-truth}};
\node at (\xstart, \ystart+\yoffset) {\includegraphics[width=\imgwidth]{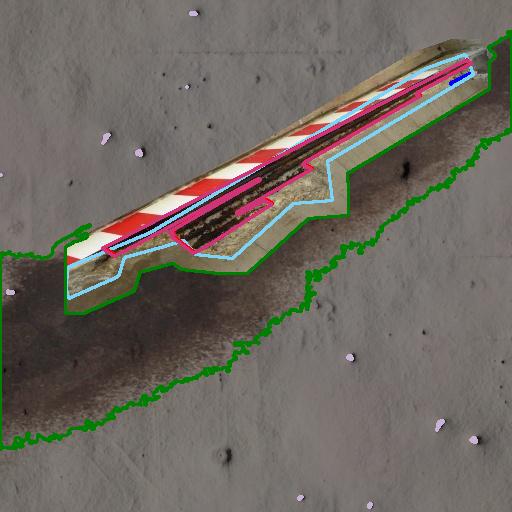}};
\node at (\xstart+\xoffset, \ystart+\yoffset) {\includegraphics[width=\imgwidth]{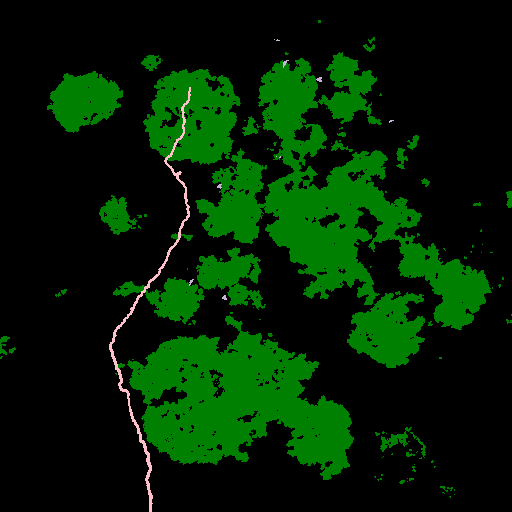}};
\node at (\xstart+\xoffset*2, \ystart+\yoffset) {\includegraphics[width=\imgwidth]{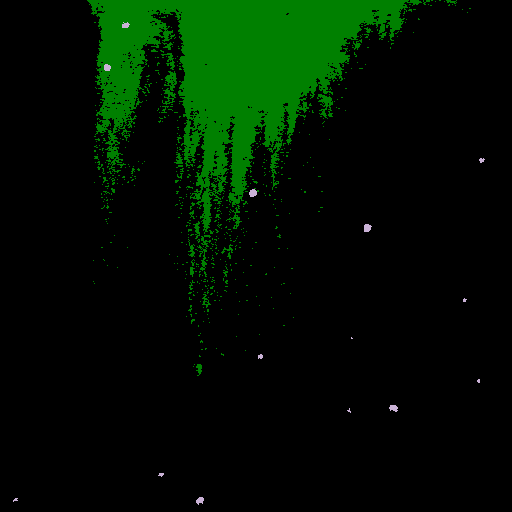}};
\node at (\xstart+\xoffset*3, \ystart+\yoffset) {\includegraphics[width=\imgwidth]{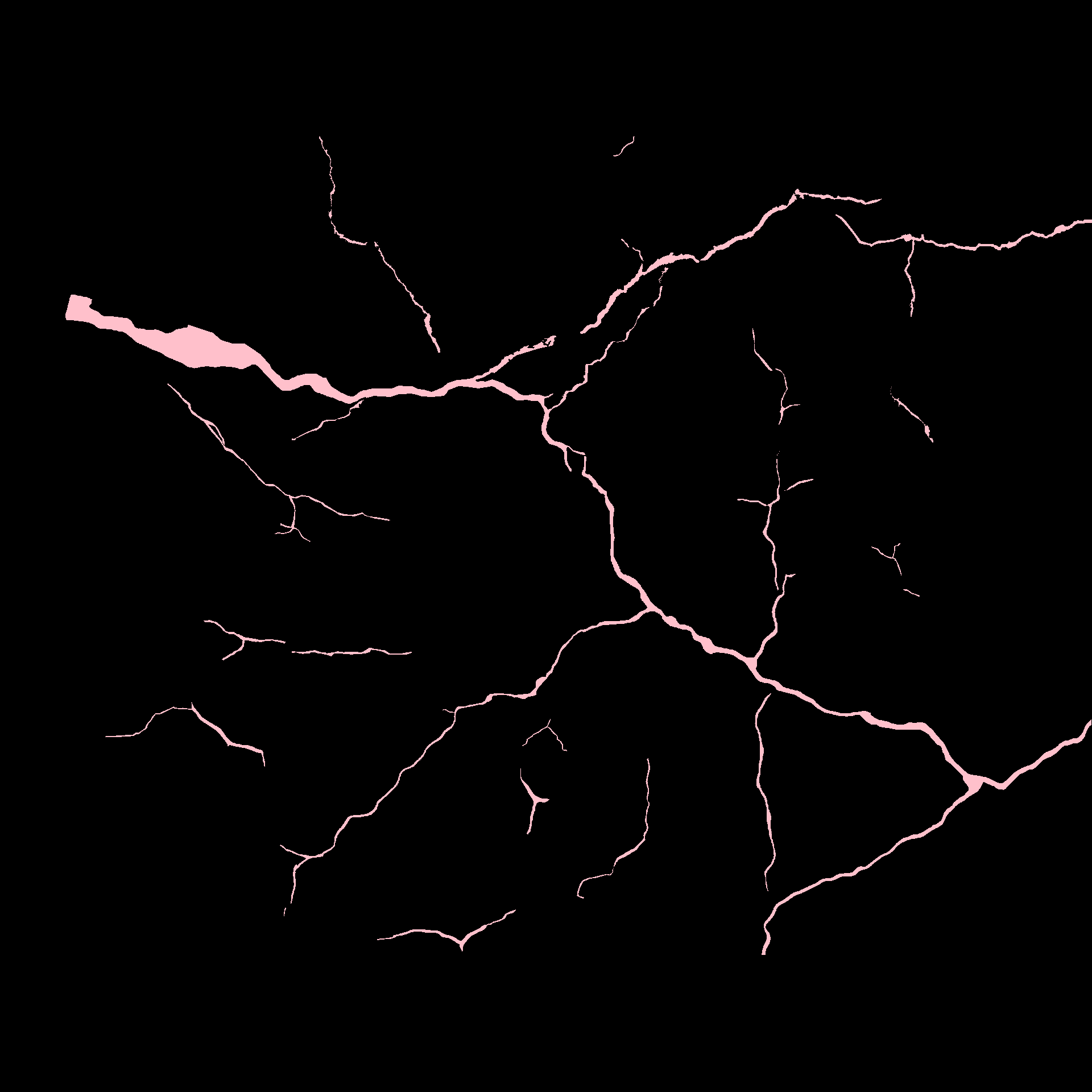}}; % new finecrack ground-truth

% Labels for the columns (top row)
\node at (\xstart, \ystart+\labeloffset) {\small daclonsynth};
\node at (\xstart+\xoffset, \ystart+\labeloffset) {\small synthcrack};
\node at (\xstart+\xoffset*2, \ystart+\labeloffset) {\small synthcavity};
\node at (\xstart+\xoffset*3, \ystart+\labeloffset) {\small finecrack}; % new label
\end{tikzpicture}

}

\caption{Left column/daclonsynth: Cropped defect shape 
from dacl10k pasted on synthetic concrete background and ground-truth below showing polygonal annotations of real-world (\textit{Spalling} with corroded \textit{Exposed Rebars}) and synthetic defects which are namely \textit{Cavity} (pale purple) and  \textit{Weathering} (green);
Middle-left Column/synthcrack: synthetic concrete surface with synthetic defects and ground-truth showing: \textit{Crack} (rose), \textit{Cavity}, \textit{Weathering};
Middle-right column/synthcavity: synthcavity sample and ground-truth showing three combined masks, \textit{Cavity} from both rendered concrete texture and the cavity generative model, and synthetic \textit{Weathering};
Right column/finecrack: Test sample from dacl10k and fine-resolution crack masks.}

\label{fig:synth-overview}
\end{figure*}

Worldwide, many bridges are exposed to high traffic loads, extreme weather events, sea salt, and de-icing chemicals, leading to defects. At the same time, most industrialized countries face a growing stock of old infrastructure \cite{artbridge2024,brueckstat}. To determine rehabilitation measurements and immediate actions, such as traffic restrictions or bridge closures, defects are identified and assessed during inspections. However, the current inspection process is often inefficient and prone to error \cite{Phares2004,LIU2023100167,abdallah2021bridge}, highlighting the significant potential of automated inspection methods to improve accuracy and reliability.
Within an automated inspection workflow semantic segmentation plays a central role as it classifies, measures and localizes damage at pixel level~\cite{dacl10k,benz2022image,HE2024102586}. 

The largest and most diverse dataset for the segmentation of bridge defects and components is dacl10k~\cite{dacl10k} due to its variety of buildings and classes.
However, a major challenge of dacl10k is the strong class imbalance, affecting not only the number of images but also the pixel-level and instance-level distributions. \Eg, the dataset contains approximately ten times more images labeled with \textit{Spalling} than with \textit{Rockpocket}, and about ten times more pixels annotated as \textit{Protective Equipment} than as \textit{Crack}. This imbalance is evident in the dacl10k challenge~\cite{dacl_challenge}, where participants consistently reported significantly poor performance for two concrete defects, \textit{crack} and \textit{cavity}.
%Fabians alt:
Addressing this imbalance is critical to improving model robustness, particularly for real-world bridge inspections, where models must be resilient to variations in image quality, camera pose, concrete texture, and the degree of weathering. Real-world applications require models to perform reliably under varying conditions, but despite the wide range of image acquisition scenarios encountered in civil engineering, no study has rigorously investigated the robustness of multi-class or multi-label computer vision models for damage detection in this domain.

In our work, we address these challenges by introducing three synthetic dataset extensions, collectively referred to as ``\textit{synth-dacl}'', each designed with 5,000 samples. 
The first extension addresses the issue of class imbalance. This enhancement superimposes real-world damage polygons on synthetic concrete backgrounds to maintain realistic defect shapes while balancing the representation of underrepresented defect classes.
The second and third extension focus on improving model performance for challenging single defect types in practice. We simulate concrete surfaces with one primary synthetic defect -- either a crack or a cavity -- per set to directly target the detection of these classes.
We systematically evaluate how synth-dacl extensions affect average performance, individual class performance, and overall model robustness. Robustness is assessed by applying 15 image perturbations to real-world test data from the dacl10k dataset, simulating conditions such as changes in illumination, noise, and contrast. This step ensures that our models are not only accurate, but also resilient to the unpredictable conditions found in real-world bridge inspections.

%%%%%%%%% RELATED WORK
\section{Related Work}
\subsection{Bridge Inspection Datasets}
The S2DS dataset~\cite{benz2022image}, which comprises 743 samples, is the first real-world semantic segmentation dataset for bridge inspection with pixel-wise labels for six classes relevant for concrete bridge inspections.
In the field of binary crack segmentation both OmniCrack30k~\cite{Benz_2024_CVPR} and CrackSeg9k~\cite{kulkarni2022crackseg9k} are dataset collections for cracked and uncracked surfaces of various materials. 

On the other hand, synthetic data has become widely used to enhance performance on real-world tasks, particularly when there is a shortage of well-labeled images \cite{poucin2021boosting}. For example, Dwibedi et al.~\cite{dwibedi2017cut} generate synthetic images by cutting and pasting object instances into diverse environments. Other studies explore the role of synthetic data in improving robustness in medical imaging. For instance, Al Khalil et al~\cite{al2023usability} examine the usability of synthesized short-axis Cardiac Magnetic Resonance (CMR) images generated using Generative Adversarial Networks to ameliorate the robustness of heart cavity segmentation models across various conditions. A comprehensive review of such applications can be found in \cite{man2022review}.

In civil engineering, synthetic data starts to become a resource to mitigate the limited availability of pixel-accurate annotated datasets. Much of the research in this field focuses on crack segmentation \cite{Jaziri_2024_WACV,xu2023innovative,SuperviselySyntheticCrackSegmentation}. For instance, \cite{Jaziri_2024_WACV} developed a simulation model in Blender for generating additional synthetic crack data for concrete surfaces. The synthetic cracks in this dataset are modeled using irregular fractals.
Another synthetic crack dataset is the Supervisely Synthetic Crack Segmentation dataset \cite{SuperviselySyntheticCrackSegmentation} which consists of 1,558 synthetic images for road surface crack detection. This dataset employs various generative algorithms, such as random walk, rapidly exploring random trees, and L-systems, to produce a diverse array of crack patterns. \cite{xu2023innovative} introduced a synthetic dataset specifically for dam crack detection. This dataset integrates crack patterns extracted from real-world, open-source datasets with a 3D mesh model of an actual dam, creating realistic training data suited to the unique structural context of dams.

While these datasets have advanced research in the field of automated bridge inspections, each faces limitations when applied to practical scenarios.
They typically focus on a single defect class (\textit{Crack}),
offer limited diversity in image quality, concrete texture, and environmental conditions, and lack lack multi-label annotations which is due to the overlapping character of concrete defects crucial.
Moreover, existing studies on the robustness of semantic damage segmentation models~\cite{ASADISHAMSABADI2022111590,lee2019robust} are, again, restricted to binary crack detection and assess only a narrow set of perturbations, which fall short of representing the wide range of real-world challenges encountered during bridge inspections.

%%%%%%%%% DACL10K
\section{Datasets}
\label{sec:datasets}
In the following, detailed information on the investigated real-world dataset dacl10k and its finecrack masks (Section~\ref{sec:dacl10k}) as well as the synthetic extensions (Section~\ref{sec:synthaddons}) are provided. 

\subsection{dacl10k}
\label{sec:dacl10k}
\textbf{dacl10k}~\cite{dacl10k} is the first large-scale dataset for automated bridge inspections, containing 9,920 polygon-annotated images across 19 classes, grouped into defects and structural components. It features multi-label semantic segmentation with coarse pixel-level annotations, which stem from real bridge inspections. The dataset includes common defect combinations, as shown in the top-left tile of Figure~\ref{fig:synth-overview}. For the underlying work a new version v3 of this dataset~\cite{RQUOYN_2025} (referred to as dacl10k) was developed. During transition class ambiguities were resolved: \textit{Spalling} and \textit{Rockpocket} were often confused in v2 due to their visual similarity, even though they arise from different causes: corroding reinforcement versus poor concrete deaeration. The same applied for \textit{Joint Tape} vs. \textit{Restformwork}, and \textit{Weathering} vs. \textit{Wetspot}. 

\subsubsection{Finecrack Masks}
\label{sec:finecrackdataset}
Most open-source datasets for crack or defect segmentation~\cite{kulkarni2022crackseg9k,Xu2019,Dorafshan2018Dec,ZOU2012227,10.1007/978-981-15-9343-7_36,LIU2019139} focus solely on cracks on plain concrete surfaces. These datasets lack important real-world variations, such as wet cracks, efflorescence, graffiti, and weathered backgrounds. Consequently, models trained on these datasets~\cite{Li_2023,10222276} tend to generate numerous false positives, as evidenced by the low precision scores in Table~\ref{tab:iou-finecr}.

From a civil engineer's perspective, cracks are not necessarily severe. In concrete structures, they occur where the tensile strength of the concrete is exceeded. Depending on the exposure of the building part, cracks with a width up to 0.4 mm (0.016 in) on non-prestressed building parts may be irrelevant regarding structural integrity and durability. However, on pre-stressed bridges, a crack width of 0.1 mm (0.004 in) can indicate tendon failure and be critical~\cite{CrackwidthDIN,ACICommittee224Cracking}. According to many inspection guidelines, the required measurement accuracy for crack width is 0.1 mm~\cite{DIN_1076,RIEBW,ACICommittee201}. This emphasizes that, for practical use, the crack defect class requires pixel-accurate segmentation.

To address this issue, we created fine-resolution crack masks for the 496 dacl10k test images that contain \textit{Crack} and \textit{ACrack}. First, the polygon is cropped and contrast-enhanced. Then, it is segmented. For Crack instances, we apply Multi-Otsu thresholding to grayscale images. For \textit{ACracks}, we use a pre-trained crack segmentation model~\cite{Li_2023} to generate approximations of fine crack masks. The results are fused into a binary mask and manually refined by a civil engineer to ensure pixel-level accuracy. Thus, the defect classes \textit{Crack} and \textit{ACrack} from dacl10k are fused within one binary \textit{Crack} mask.

\subsection{Synthetic Dataset Extensions}
\label{sec:synthaddons}
Due to the significant class imbalance and low performance on specific classes in the dacl10k dataset, as well as the high costs associated with data collection and labeling, we explore synthetic data generation methods in the following sections \cite{9D6E4M_2025}. Details on the class distribution in the original dacl10k-v3 dataset are provided in our supplementary material.

\subsubsection{Synthetic Concrete Surfaces}
To increase robustness in defect segmentation and overcome class imbalance, we introduce three new dataset extensions based on synthetic concrete surfaces: \textbf{daclonsynth}, \textbf{synthcrack} and \textbf{synthcavity}. To generate these surfaces an extended version of the physics-based rendering (PBR) introduced in Jaziri~\etal~\cite{Jaziri_2024_WACV} was used.
The rendering pipeline consists of two main stages: (1) scene generation, and (2) defect injection. In the first stage, various texture maps are applied to produce diverse concrete surfaces using Blender’s Cycles PBR engine. Optional overlays such as moss or dirt simulate \textit{Weathering} effects (see Figure~\ref{fig:synth-overview}). In the second stage, up to two defects are added per scene (see Section~\ref{sec:synthcrack} and~\ref{sec:synthcavity}), and semantic ground-truths including depth, surface normals, and class masks, are generated automatically, including a dedicated map for \textit{Weathering}.

\subsubsection{daclonsynth} 

\begin{figure*}
    \centering
    \includegraphics[width=\linewidth]{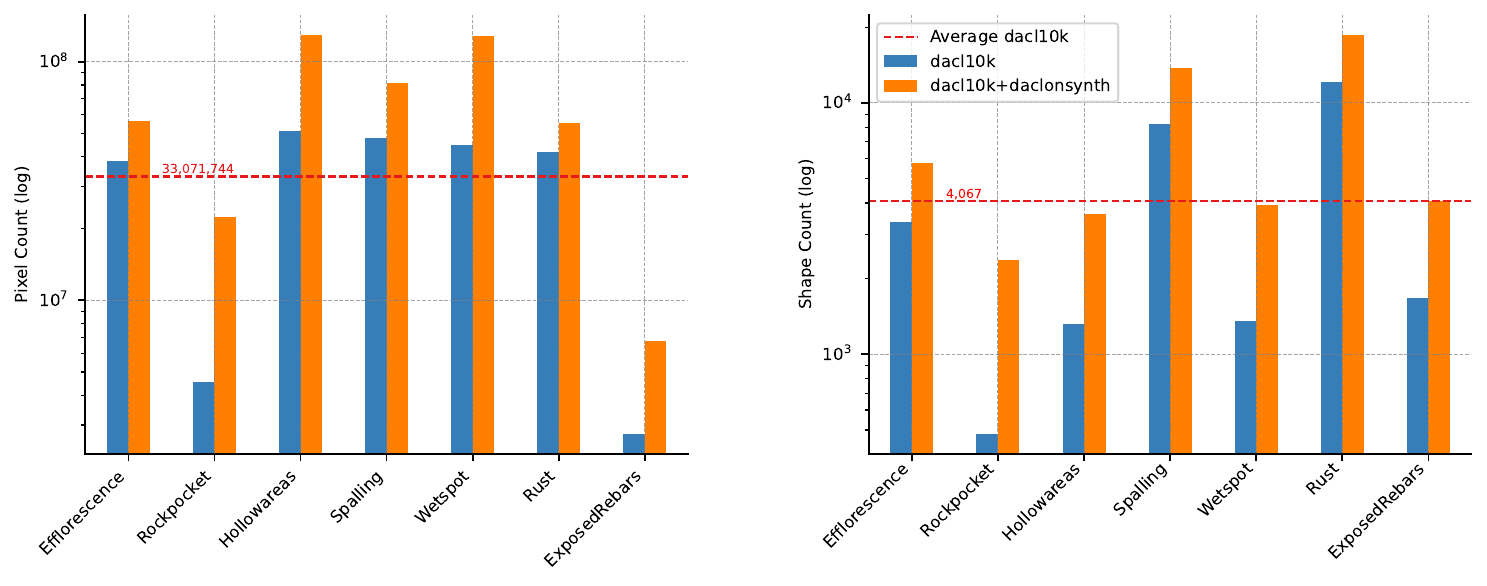}
    \caption{Comparison of classwise pixel count (left) and shape count (right) of dacl10k train set (blue) and the combination of dacl10k and daclonsynth (orange). The red dashed line marks the average pixel, or rather shape count, over the displayed classes from dacl10k. All stats based on the resized data ($512\times512$).}
\label{fig:daclonsynth-barchart}
\end{figure*}

Building on these synthetic scenes, we generate the \textbf{daclonsynth} extension with the specific goal of mitigating class imbalance in the dacl10k training set. As shown in Figure~\ref{fig:daclonsynth-barchart}, several classes, particularly \textit{Rockpocket}, \textit{Exposed Rebars}, \textit{Hollowareas}, and \textit{Wetspot}, are heavily underrepresented in both pixel count and shape count. For instance, \textit{Rockpocket} is annotated only 354 times and accounts for merely 4.5 million pixels, while the average across classes is around 45 million pixels.

To generate new training samples, we first filter the dacl10k training images to include only those containing at least one instance of a targeted underrepresented class. Then, each annotated shape is cropped from the image along with its corresponding polygonal annotation. This cropped region is then randomly rotated and pasted onto a synthetic concrete surface that is also randomly selected. For each class, half of the synthetic samples include \textit{Weathering} overlays, while the other half remain clean to promote better generalization to both conditions.

In cases where the target defect typically co-occurs with a larger structural issue, as is common for \textit{Exposed Rebars} appearing within \textit{Spalling} or \textit{Rockpocket}, the crop is extended to include the full area of the co-located host defect. Furthermore, to prevent models from exploiting the artificial distinction between real annotations and synthetic backgrounds, the shapes of \textit{Spalling}, \textit{Rockpocket}, \textit{Wetspot}, \textit{Hollowareas}, and \textit{Efflorescence} are dilated using a $30\times30$ kernel prior to compositing.

To determine the number of synthetic samples needed per class, we calculate the number of instances required to bring the pixel and shape counts closer to the averages of all underrepresented classes. We then use the mean of these two estimates to define a target sample count per class. The final distribution of the 5,000 synthetic samples in \textbf{daclonsynth} is determined proportionally based on these class-wise demands while accounting for the average number of pixels and shapes per sample for each defect. Although complete balance is not possible due to defect co-occurrence and overlaps, the resulting dataset significantly shifts the class distribution towards uniformity. Non-underrepresented classes may also be reproduced due to overlaps, which further contributes to a more diverse and realistic training dataset.

\subsubsection{synthcrack}
\label{sec:synthcrack}
Crack patterns are generated using a fractal model based on~\cite{Jaziri_2024_WACV} and rendered onto synthetic concrete surfaces with corresponding semantic masks. The resulting extension, \textbf{synthcrack} matches dacl10k in terms of crack pixel and shape counts. It also contains some incidental \textit{Cavity} shapes, originating from fine-grained surface geometry embedded in the rendering pipeline, but these are not generated explicitly.

\subsubsection{synthcavity}
\label{sec:synthcavity}
To generate cavities, we introduce a dedicated simulation approach based on Perlin noise. Multiple noise layers with varying octaves (2–32), persistence (0.6–0.9), and lacunarity (1.5 or 2) are combined to create irregular cavity maps. After thresholding and filtering out small regions, the resulting masks are used to build geometry-aware PBR textures. These are rendered and overlaid onto synthetic concrete surfaces to form the synthcavity dataset. Compared to dacl10k, synthcavity contains more cavity shapes, but with smaller area per instance, leading to a lower overall pixel count.

\begin{figure*}
    \centering
    \includegraphics[width=\linewidth]{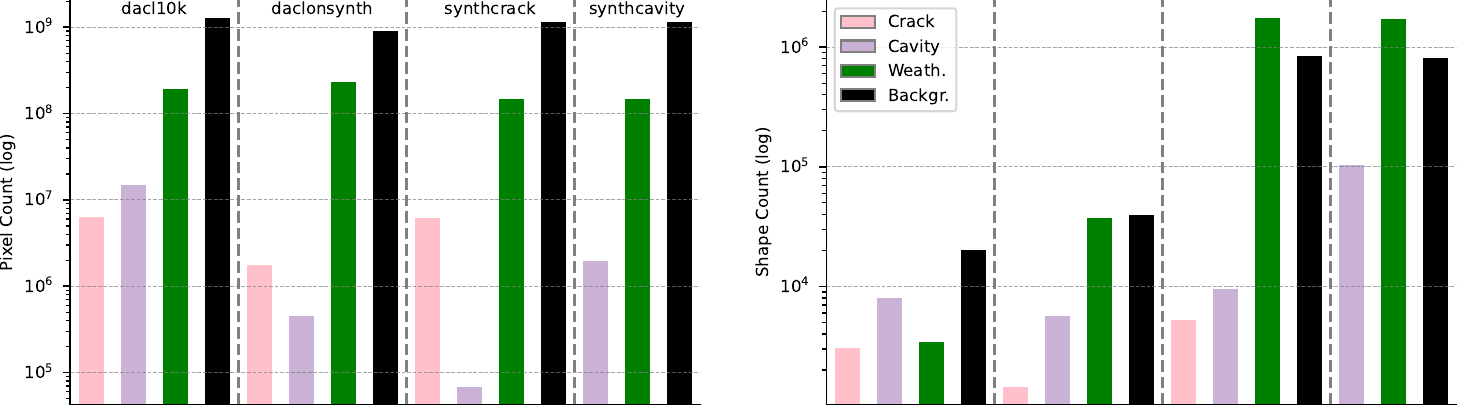}
\caption{Pixel (left) and shape (right) counts for \textit{Crack}, \textit{Cavity}, \textit{Weathering}, and background in dacl10k and synthetic datasets.}
\label{fig:counts-added}
\end{figure*}

All synthetic datasets also contribute additional background and \textit{Weathering} annotations. Overlaps in \textbf{daclonsynth} further introduce crack and cavity shapes, which are reflected in the distributions shown in Figure~\ref{fig:counts-added}.

% \begin{wrapfigure}{r}{0.5\linewidth}
% \centering
% \includegraphics[width=\linewidth]{img/Figures_final/combined_classw_final.pdf}
% \caption{Pixel (left) and shape (right) counts for \textit{Crack}, \textit{Cavity}, \textit{Weathering}, and background in dacl10k and synthetic datasets.}
% \label{fig:counts-added}
% \vspace{-2em}
% \end{wrapfigure}

%%%%%%%%% BASELINES
\section{Experiments}
\label{sec:experiments}

In this section, we evaluate the effectiveness of our synthetic dataset extensions by testing eight differently trained models on the dacl10k test set. To assess the contribution of synthetic data to model robustness, the same models are also evaluated on a perturbed version of the test set. Additionally, we conduct two ablation studies: one focusing on the segmentation performance for finely annotated cracks, and another analyzing cross-domain generalization from synthetic to real-world data.

All experiments employ a Feature Pyramid Network (FPN) \cite{Lin_2017_CVPR} with a MaxViT-Base Vision Transformer backbone \cite{maxvit}. The primary evaluation metric is Intersection over Union (IoU) as previous work~\cite{Everingham2014ThePV,7780719,8100027} where IoU is set to 1 when the union is zero. This is complemented by F1 score, Precision, and Recall the analysis of the fine-grained classes and the perturbed testing. In general, the metrics are computed per class at the image level and averaged across the dataset. Mean values are reported by averaging the class-level scores.
Further details, such as split size and class imbalance, can be found in the supplementary material.

\subsection{Results on dacl10k}
\label{sec:res-on-dacl}

As shown in Table~\ref{tab:iou}, the inclusion of synthetic data improves overall performance, with the six highest mIoU results being achieved when synthetic data is used during training. At class level, extending dacl10k with the semi-synthetic daclonsynth improves IoU in average for the daclonsynth classes by 1.3\%, with \textit{Rockpocket} experiencing a notable +4\% increase (see Table~\ref{tab:iou}). This boost likely stems from the fivefold increase in representation of this class provided by daclonsynth. 
Notably, the top three configurations for crack IoU in Table~\ref{tab:iou} do not include synthcrack, which might seem contradictory at first. This outcome is explained by the fact that synthcrack employs much finer and more detailed crack annotations than those in dacl10k. As a result, the model learns to predict cracks at a finer resolution than what the dacl10k ground-truth can capture. This hypothesis is supported by the result of the model trained only on synthcrack, presented in Table~\ref{tab:CrossVal}, which confirms best performance on the finecrack masks.
The results on the finecrack masks, and on \textit{Cavity} are analyzed in the following Chapter~\ref{sec:fine-grained}. 

\begin{table*}
\tiny
\centering
\caption{\textbf{mIoU and classwise IoU on the dacl10k test split.} 
A check mark in the Training Data column indicates the synthetic data used during training. Bold numbers indicate the highest value in each column, while underlined numbers represent the lowest.}
\label{tab:iou}
\resizebox{\textwidth}{!}{%

\begin{tabular}{cccccccccccccccccccccccc}
\toprule
\multicolumn{4}{c}{\textbf{Train data}} &  &\multicolumn{19}{c}{\textbf{classwise IoU}}\\ \cmidrule(r){1-4}\cmidrule(l){6-24}
\rotatebox{90}{dacl10k} & \rotatebox{90}{daclonsynth} & \rotatebox{90}{synthcrack} & \rotatebox{90}{synthcavity} & \rotatebox{90}{\textbf{mIoU}} & \rotatebox{90}{Crack} & \rotatebox{90}{ACrack} & \rotatebox{90}{Effloresc.} & \rotatebox{90}{Rockpocket} & \rotatebox{90}{WConccor} & \rotatebox{90}{Hollowareas} & \rotatebox{90}{Cavity} & \rotatebox{90}{Spalling} & \rotatebox{90}{Restformw.} & \rotatebox{90}{Wetspot} & \rotatebox{90}{Rust} & \rotatebox{90}{Graffiti} & \rotatebox{90}{Weathering} & \rotatebox{90}{ExposedR.} & \rotatebox{90}{Bearing} & \rotatebox{90}{EJoint} & \rotatebox{90}{Drainage} & \rotatebox{90}{PEquipment} & \rotatebox{90}{JTape} \\
\midrule
\checkmark & \xmark & \xmark & \xmark & 42.71 & 31.40 & 49.81 & 42.97 & 21.26 & \underline{9.79} & 54.88 & 20.07 & \underline{44.83} & 31.69 & 23.52 & \underline{47.03} & 66.09 & \textbf{37.26} & 39.89 & 59.16 & 56.44 & 60.17 & 73.61 & 41.64 \\
\checkmark & \checkmark & \xmark & \xmark & 42.93 & \textbf{31.42} & \textbf{51.76} & \textbf{43.53} & \textbf{25.26} & 10.13 & 56.07 & 22.02 & 45.19 & 31.68 & 24.22 & 48.53 & 66.01 & \underline{27.11} & 40.61 & 59.47 & 55.12 & \textbf{61.62} & \underline{73.00} & 42.85 \\
\checkmark & \checkmark & \checkmark & \xmark & 43.12 & 30.16 & 49.76 & 42.73 & 22.64 & \textbf{14.25} & 55.62 & \underline{17.58} & 45.50 & 32.46 & \underline{18.43} & 48.04 & 66.00 & 36.89 & \underline{39.59} & \textbf{60.98} & \textbf{59.96} & 60.55 & 73.20 & \textbf{44.97} \\
\checkmark & \checkmark & \xmark & \checkmark & 42.85 & 30.62 & 47.05 & 42.58 & 22.84 & 13.01 & 56.41 & 20.95 & 45.68 & 32.28 & 22.91 & 48.20 & \underline{63.63} & 34.82 & 40.39 & 58.44 & 56.22 & 61.26 & 74.23 & 42.58 \\
\checkmark & \checkmark & \checkmark & \checkmark & \textbf{43.37} & 29.48 & 50.55 & 43.25 & 21.11 & 11.99 & \textbf{56.90} & 17.89 & 45.65 & \textbf{32.60} & 23.87 & \textbf{49.70} & 65.65 & 35.87 & \textbf{41.73} & 60.92 & 58.56 & 60.73 & 74.10 & 43.51 \\
\checkmark & \xmark & \checkmark & \xmark & 42.76 & \underline{27.25} & 48.55 & 43.23 & 22.47 & 11.14 & 55.18 & 23.14 & \textbf{46.08} & \underline{31.34} & 24.62 & 48.45 & 66.18 & 32.17 & 40.96 & 60.50 & \underline{54.86} & 59.47 & \textbf{74.70} & 42.14 \\
\checkmark & \xmark & \xmark & \checkmark & 42.88 & 29.18 & 47.93 & 43.15 & 22.86 & 11.49 & 55.02 & \textbf{24.63} & 45.44 & 31.74 & \textbf{24.77} & 48.57 & \textbf{66.51} & 33.26 & 40.03 & 60.28 & 55.77 & 58.58 & 74.00 & 41.46 \\
\checkmark & \xmark & \checkmark & \checkmark & \underline{41.86} & 30.68 & \underline{46.82} & \underline{42.41} & \underline{20.21} & 11.55 & \underline{54.50} & 19.06 & 45.08 & 32.00 & 22.61 & 48.30 & 64.46 & 36.98 & 39.61 & \underline{54.38} & 56.30 & \underline{56.50} & 73.05 & \underline{40.84} \\
\bottomrule
\end{tabular}
}
\end{table*}
% \endgroup

\subsection{Ablation: Results on Fine-Grained Classes}
\label{sec:fine-grained}

For both fine-grained classes, \textit{Crack} in the form of finecrack masks and \textit{Cavity}, we provide additional metrics in Table~\ref{tab:iou-finecr}.
All models incorporating synthcrack report better results on finecrack masks than the baseline trained on dacl10k only. The best finecrack IoU is achieved by the model utilizing dacl10k, daclonsynth and synthcrack, 1.3 percent points higher than dacl10k. 
Regarding \textit{Cavity} the highest IoU (24.63\%) and F1 score (39.53\%) is reported for the model trained on the combination of dacl10k and synthcavity.

\begingroup
\setlength{\tabcolsep}{4pt} % Default value: 6pt
\renewcommand{\arraystretch}{.9} % Default value: 1
\begin{table}
\tiny
\centering
\caption{\textbf{IoU, F1 Score, Recall and Precision on the finecrack and cavity masks.} 
Bold numbers indicate the highest value in each column, while underlined numbers represent the lowest. We compare with two open-source baseline methods for crack segmentation (two bottom rows). The train data are: (1) dacl10k, (2) daclonsynth, (3) synthcrack, (4) synthcavity.}
\label{tab:iou-finecr}
\begin{tabular}{cccccccccccc}
\toprule
\multicolumn{4}{c}{\textbf{Train data}} & \multicolumn{4}{c}{\textbf{Metrics Finecrack}} & \multicolumn{4}{c}{\textbf{Metrics Cavity}}\\
\cmidrule(r){1-4} \cmidrule(lr){5-8} \cmidrule(l){9-12}
(1) & (2) & (3) & (4) & IoU & F1 & Rec. & Prec. & IoU & F1 & Rec. & Prec.\\
\midrule
\checkmark & \xmark & \xmark & \xmark  & 11.54 & 20.69 & 28.87 &  16.13 & 20.07 & 33.43 & 31.06 & \underline{36.20}\\
\checkmark & \checkmark & \xmark & \xmark  & 12.09 & 21.57 & 31.19 & 16.48 & 22.02 & 36.09 & 33.68 & 38.87 \\
\checkmark & \checkmark & \checkmark & \xmark & \textbf{12.88} & \textbf{22.82} & 26.64 & 19.96 & \underline{17.58} & \underline{29.91} & 22.34 & 45.23\\
\checkmark & \checkmark & \xmark & \checkmark & 11.74 & 21.02 & 31.99 & 15.65 & 20.95 & 34.65 & 26.85 & 48.83 \\
\checkmark & \checkmark & \checkmark & \checkmark & 11.97 & 21.38 & 23.86 & 19.37 & 17.89 & 30.35 & \underline{21.46} & \textbf{51.82}\\
\checkmark & \xmark & \checkmark & \xmark & 12.12 & 21.38 & \underline{22.17} & \textbf{21.11} & 23.14 & 37.59 & \textbf{36.11} & 39.20\\
\checkmark & \xmark & \xmark & \checkmark & \underline{11.34} & \underline{20.37} & \textbf{34.11} & \underline{14.52} & \textbf{24.63} & \textbf{39.53} & 33.86 & 47.48\\
\checkmark & \xmark & \checkmark & \checkmark & 12.33 & 21.95 & 26.50 & 18.73 & 19.06 & 32.02 & 23.20 & 51.66 \\

\midrule
\multicolumn{4}{l}{HrSegNet-B64~\cite{Li_2023}} & 3.34 & 6.47 & 19.26 & 3.89 & \xmark & \xmark & \xmark & \xmark\\
\multicolumn{4}{l}{CT-CrackSeg~\cite{10222276}} & 0.59 & 1.18 & 84.25 & 0.60 & \xmark & \xmark & \xmark & \xmark \\
\bottomrule
\end{tabular}
\end{table}
\endgroup

\subsection{Results on perturbed dacl10k}
In real-world bridge inspection applications, factors such as different camera models, varying image acquisition settings and environmental conditions introduce noise that negatively affect model performance. 
To investigate how this affects our differently trained models, we follow the methodology of Wang et al.~\cite{WANG2024110685} and apply 15 different perturbations to the test images. These include different noise functions, blur, brightness changes, and weather effects, which are demonstrated in the supplementary material. The results, averaged over all perturbations, are presented in Table~\ref{tab:testall_pert}.

Apart from the model trained on dacl10k, synthcrack and synthcavity (see Section~\ref{sec:limits}) all models trained on both original and synthetically generated data consistently show the ``most robust'' to these perturbations.
The highest IoU, F1 score, and Precision is reported for the model trained on dacl10k combined with all \textit{synth-dacl} extensions. Furthermore, this model shows 1.89\% less relative performance loss in mean IoU and 2.54\% in mean F1 score (see Table~\ref{tab:testall_pert}).

\begin{table*}
\tiny
\centering 
\caption{\textbf{Comparison of IoU, F1, Recall, and Precision on the test split between raw and perturbed (Pert.) images.} The \textit{Change} column represents the relative performance difference between testing on raw images and perturbed images, highlighting the degradation in performance caused by the perturbations. Bold numbers indicate the highest value in each column, while underlined numbers represent the lowest. The train data are: (1) dacl10k, (2) daclonsynth, (3) synthcrack, (4) synthcavity.}
\label{tab:testall_pert}
\begin{tabular}{ccccccccccccccccccc}
\toprule
\multicolumn{4}{c}{\textbf{Train data}} & 
\multicolumn{3}{c}{\textbf{mIoU}} & 
\multicolumn{3}{c}{\textbf{mF1}} & 
\multicolumn{3}{c}{\textbf{mRecall}} & 
\multicolumn{3}{c}{\textbf{mPrecision}} \\
\cmidrule(r){1-4} \cmidrule(lr){5-7} \cmidrule(lr){8-10} \cmidrule(lr){11-13} \cmidrule(lr){14-16}
(1) & 
(2) & 
(3) & 
(4) & 
Raw & Pert. & Change & 
Raw & Pert. & Change & 
Raw & Pert. & Change & 
Raw & Pert. & Change \\
\midrule 
\checkmark & \xmark & \xmark & \xmark & 42.71 & 30.87 & -27.73 & 57.89 & 43.34 & -25.13 & 56.02 & 38.63 & -31.04 & 60.38 & 60.79 & 0.67 \\
\checkmark & \checkmark & \xmark & \xmark & 42.93 & 31.03 & -27.72 & 58.12 & 43.50 & -25.16 & 55.90 & 39.30 & -29.69 & 62.18 & 61.56 & -0.99 \\
\checkmark & \checkmark & \checkmark & \xmark & 43.12 & 30.89 & -28.36 & 58.23 & 43.34 & -25.57 & 55.42 & 38.76 & -30.07 & 63.01 & 61.95 & -1.69 \\
\checkmark & \checkmark & \xmark & \checkmark & 42.85 & 30.87 & -27.95 & 58.16 & 43.51 & -25.19 & 56.83 & 39.84 & -29.90 & 60.96 & \underline{60.61} & -0.58 \\
\checkmark & \checkmark & \checkmark & \checkmark & \textbf{43.37} & \textbf{32.16} & -25.84 & \textbf{58.47} & \textbf{45.26} & -22.59 & \underline{54.98} & 40.11 & \textbf{-27.05} & \textbf{64.20} & \textbf{63.05} & \underline{-1.79} \\
\checkmark & \xmark & \checkmark & \xmark & 42.76 & 31.14 & -27.18 & 57.98 & 43.84 & -24.39 & 56.00 & 38.90 & -30.52 & 61.35 & 61.88 & 0.86 \\
\checkmark & \xmark & \xmark & \checkmark & 42.88 & 32.00 & \textbf{-25.36} & 58.20 & 45.06 & \textbf{-22.58} & \textbf{57.79} & \textbf{41.35} & -28.45 & \underline{59.79} & 60.81 & \textbf{1.70} \\
\checkmark & \xmark & \checkmark & \checkmark & \underline{41.86} & \underline{29.96} & \underline{-28.43} & \underline{57.16} & \underline{42.42} & \underline{-25.80} & 55.59 & \underline{38.07} & \underline{-31.52} & 60.26 & 60.71 & 0.75 \\
\bottomrule
\end{tabular}
\end{table*}

\subsection{Ablation: Domain-Partitioned Evaluation}

To evaluate cross-domain generalization from synthetic to real-world data, we perform a domain-partitioned ablation in which models are trained exclusively on synthetic datasets and evaluated on the dacl10k test split, thereby isolating the transferability of synthetic feature representations (see Table~\ref{tab:CrossVal}). Our findings indicate that, with the exception of synthcavity, the \textit{synth-dacl} extensions are representative and consistent to generalize to real-world defect types. This is highlighted by the highest IoU on finecrack masks (13.11\%) reported for the model trained on dacl10k and synthcrack. 
Utilizing synthcavity leads, according to Table~\ref{tab:CrossVal}, to 0\% IoU on Cavity from dacl10k test set which is further discussed in Chapter~\ref{sec:limits}. 

\begin{table}
\tiny
\centering 
\caption{\textbf{Synthetic-only training on three datasets; evaluated on real-world dacl10k test split and finecrack masks.}}
\begin{tabular}{ccccccccccccccc}
\toprule
\multicolumn{3}{c}{\textbf{Train data}} &  &\multicolumn{9}{c}{\textbf{classwise IoU}}\\ \cmidrule(r){1-3}\cmidrule(l){5-14}
\rotatebox{90}{daclonsynth} & \rotatebox{90}{synthcrack} & \rotatebox{90}{synthcavity} & \rotatebox{90}{\textbf{mIoU}} & \rotatebox{90}{Crack} & \rotatebox{90}{Finecrack} & \rotatebox{90}{Effloresc.} & \rotatebox{90}{Rockpocket} & \rotatebox{90}{Hollowareas} & \rotatebox{90}{Cavity} & \rotatebox{90}{Spalling} & \rotatebox{90}{Wetspot} & \rotatebox{90}{Rust} & \rotatebox{90}{ExposedR.}\\
\midrule
\checkmark & \xmark & \xmark & \textbf{27.42} & \xmark & \xmark & \textbf{29.54} & 20.37 & 35.34 & \xmark & \textbf{31.81} & \textbf{3.29} & \textbf{37.15} & 34.42 \\ 
        \checkmark & \checkmark & \xmark & 26.53 & 21.04 & \textbf{13.11} & 26.82 & \textbf{21.77} & \textbf{38.34} & \xmark & 31.22 & 3.22 & 35.35 & \textbf{34.47} \\ 
        \checkmark & \checkmark & \checkmark & 22.59 & \textbf{21.76} & 12.64 & 27.72 & 14.03 & 34.85 & 0.00 & 34.71 & 3.09 & 34.14 & 32.99 \\ 
        \xmark & \checkmark & \xmark & \xmark & 6.60 & 8.29 & \xmark & \xmark & \xmark & \xmark & \xmark & \xmark & \xmark & \xmark \\ 
        \xmark & \xmark & \checkmark & \xmark & \xmark & \xmark & \xmark & \xmark & \xmark & 0.15 & \xmark & \xmark & \xmark & \xmark \\ 
     \bottomrule
    \end{tabular}
    \label{tab:CrossVal}
\end{table}

\subsection{Qualitative Evaluation}
\begin{figure}
\centering
\resizebox{\linewidth}{!}{
\begin{tikzpicture}

% Set image width and spacing
\def\imgwidth{1.94cm}
\def\xstart{2}
\def\ystart{0}
\def\xoffset{1.99}
\def\yoffset{-1.99}

% Row labels (left of first column)
\node at (\xstart-1.1, \ystart) {\tiny \rotatebox{90}{ground-truth}};
\node at (\xstart-1.1, \ystart+\yoffset) {\tiny \rotatebox{90}{dacl10k}};
\node at (\xstart-1.1, \ystart+\yoffset*2) {\tiny \rotatebox{90}{dacl10k+synth}};
\node at (\xstart-1.25, \ystart+\yoffset*3) {\tiny \rotatebox{90}{\shortstack{dacl10k\\+all synth}}};

% Column headers (top row)
\node at (\xstart, \ystart+1.1) {\tiny Efflorescence};
\node at (\xstart+\xoffset, \ystart+1.1) {\tiny Rockpocket};
\node at (\xstart+\xoffset*2, \ystart+1.1) {\tiny Wetspot};
\node at (\xstart+\xoffset*3, \ystart+1.1) {\tiny Crack};
\node at (\xstart+\xoffset*4, \ystart+1.1) {\tiny Cavity};

% Row 1: ground-truth
\node at (\xstart, \ystart) {\includegraphics[width=\imgwidth]{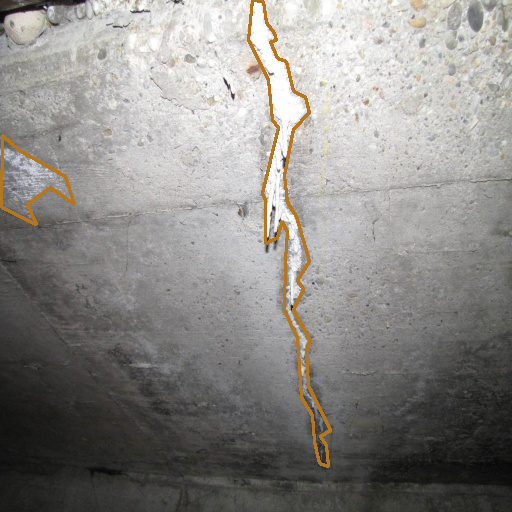}};
\node at (\xstart+\xoffset, \ystart) {\includegraphics[width=\imgwidth]{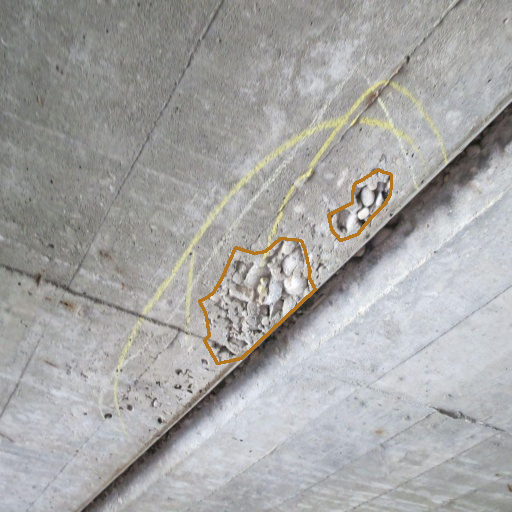}};
\node at (\xstart+\xoffset*2, \ystart) {\includegraphics[width=\imgwidth]{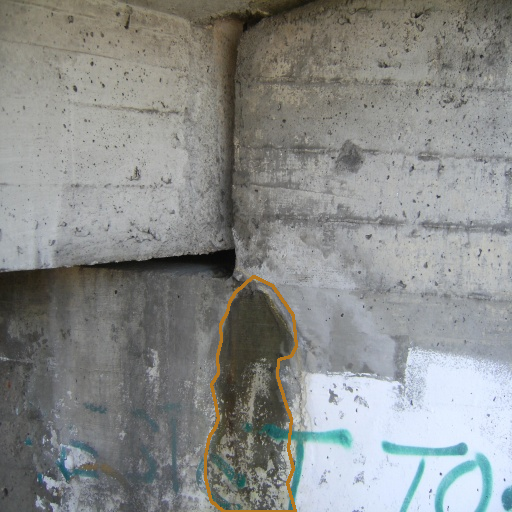}};
\node at (\xstart+\xoffset*3, \ystart) {\includegraphics[width=\imgwidth]{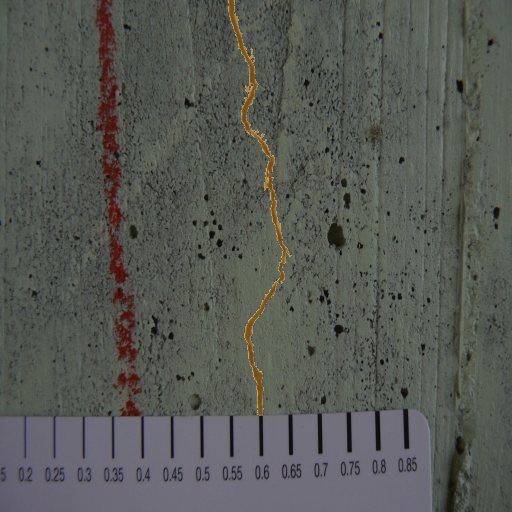}};
\node at (\xstart+\xoffset*4, \ystart) {\includegraphics[width=\imgwidth]{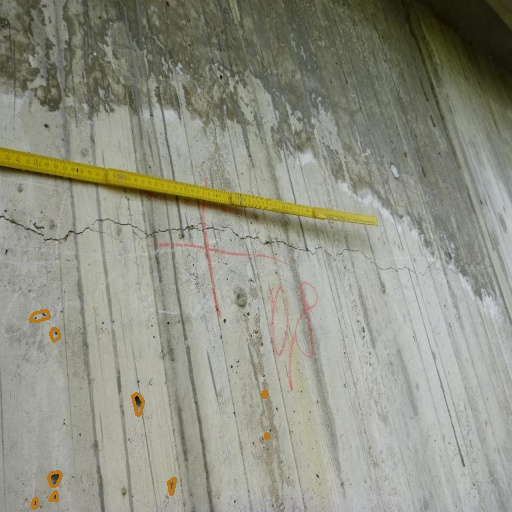}};

% Row 2: dacl10k
\node at (\xstart, \ystart+\yoffset) {\includegraphics[width=\imgwidth]{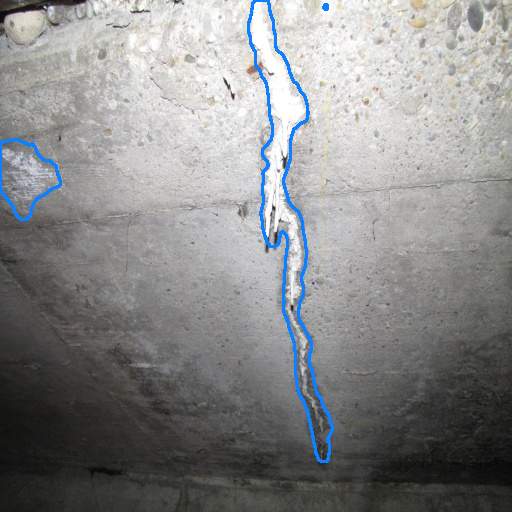}};
\node at (\xstart+\xoffset, \ystart+\yoffset) {\includegraphics[width=\imgwidth]{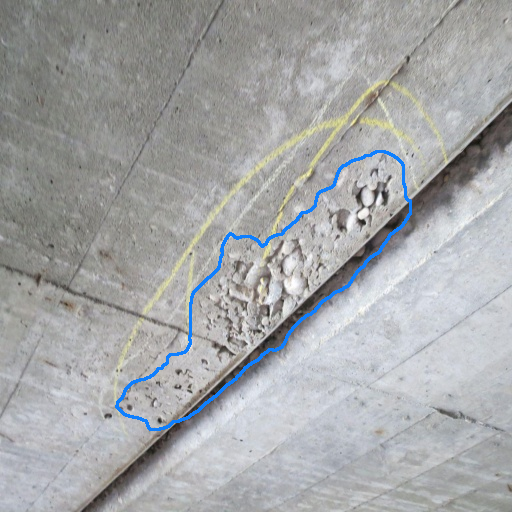}};
\node at (\xstart+\xoffset*2, \ystart+\yoffset) {\includegraphics[width=\imgwidth]{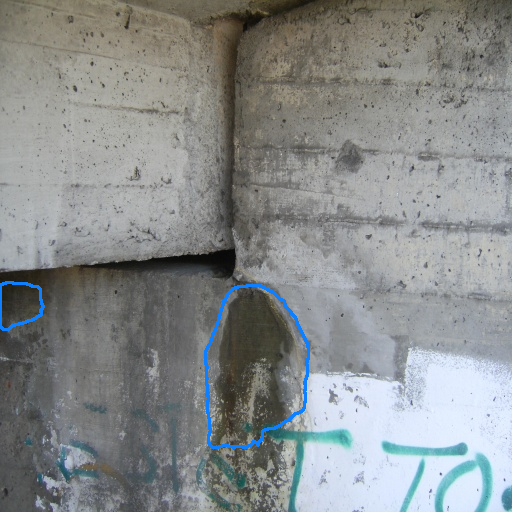}};
\node at (\xstart+\xoffset*3, \ystart+\yoffset) {\includegraphics[width=\imgwidth]{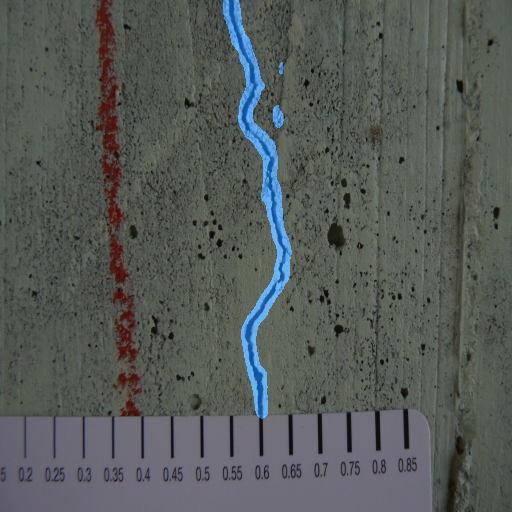}};
\node at (\xstart+\xoffset*4, \ystart+\yoffset) {\includegraphics[width=\imgwidth]{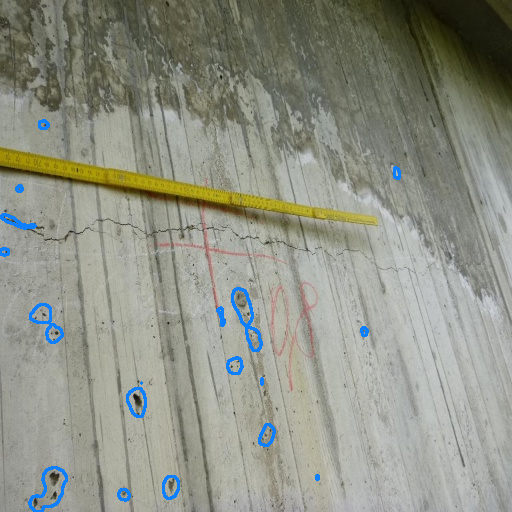}};

% Row 3: dacl10k + synth
\node at (\xstart, \ystart+\yoffset*2) {\includegraphics[width=\imgwidth]{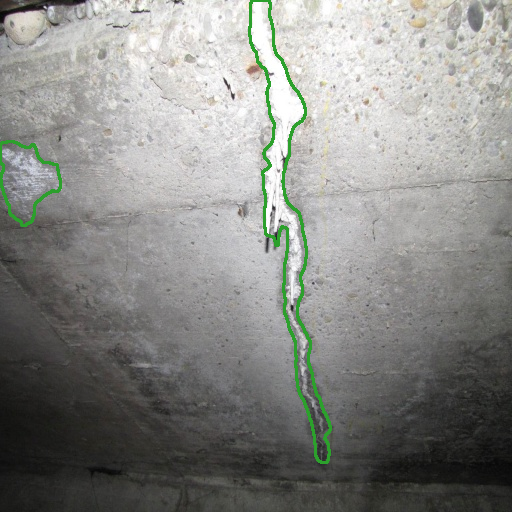}};
\node at (\xstart+\xoffset, \ystart+\yoffset*2) {\includegraphics[width=\imgwidth]{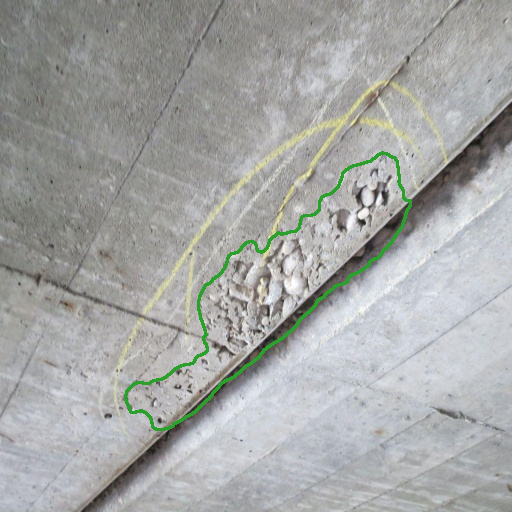}};
\node at (\xstart+\xoffset*2, \ystart+\yoffset*2) {\includegraphics[width=\imgwidth]{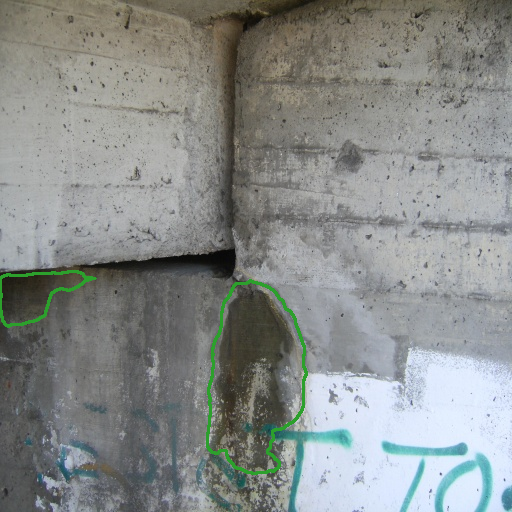}};
\node at (\xstart+\xoffset*3, \ystart+\yoffset*2) {\includegraphics[width=\imgwidth]{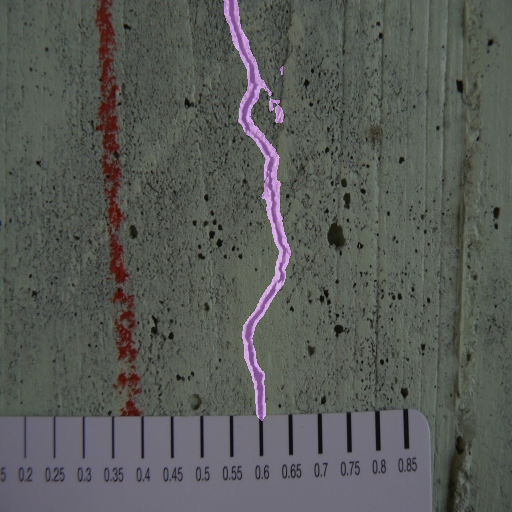}};
\node at (\xstart+\xoffset*4, \ystart+\yoffset*2) {\includegraphics[width=\imgwidth]{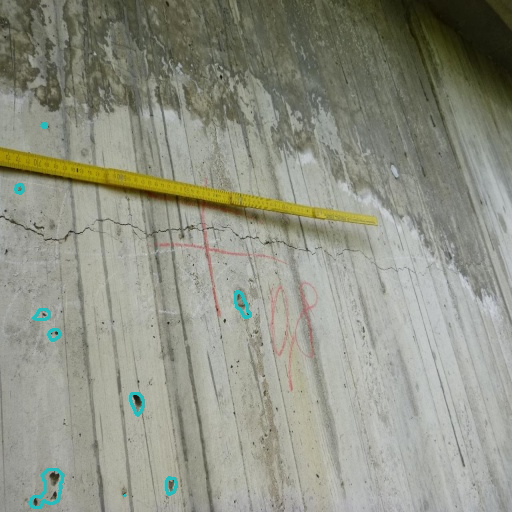}};

% Row 4: dacl10k + all synth
\node at (\xstart, \ystart+\yoffset*3) {\includegraphics[width=\imgwidth]{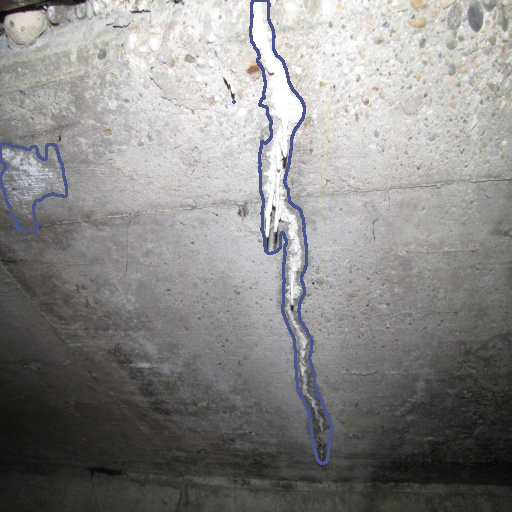}};
\node at (\xstart+\xoffset, \ystart+\yoffset*3) {\includegraphics[width=\imgwidth]{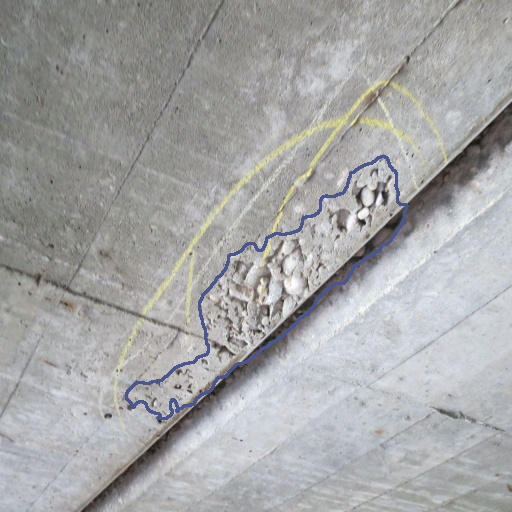}};
\node at (\xstart+\xoffset*2, \ystart+\yoffset*3) {\includegraphics[width=\imgwidth]{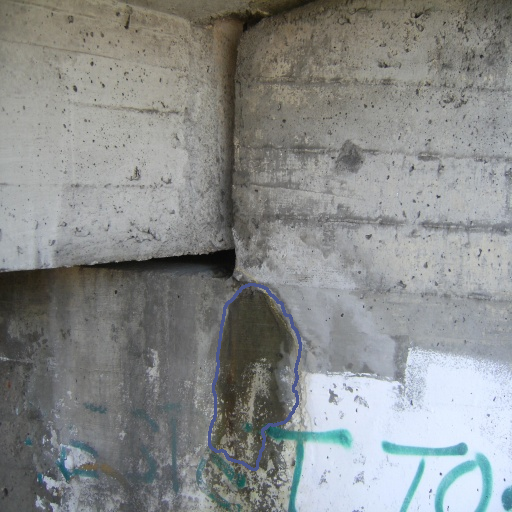}};
\node at (\xstart+\xoffset*3, \ystart+\yoffset*3) {\includegraphics[width=\imgwidth]{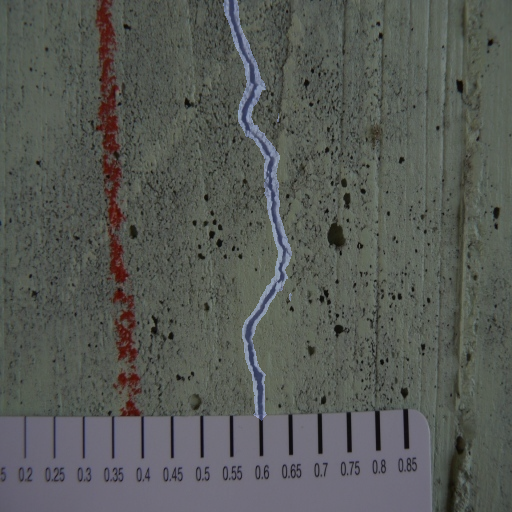}};
\node at (\xstart+\xoffset*4, \ystart+\yoffset*3) {\includegraphics[width=\imgwidth]{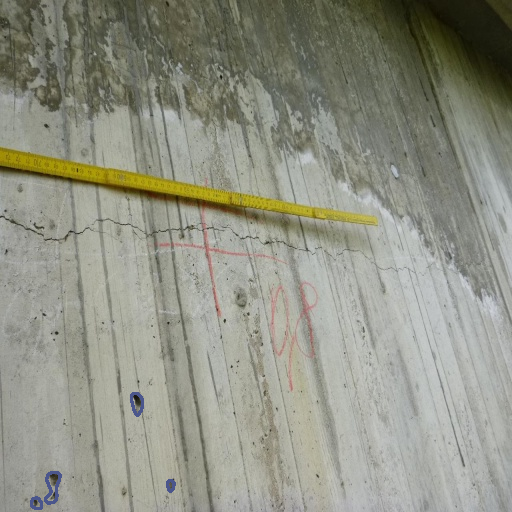}};

\end{tikzpicture}
}
\caption{Qualitative results on dacl10k test samples for different damage classes (columns) across different training data setups (rows), where the row dacl10k+synth shows the predictions of the model achieving highest IoU on the given class.}
\label{fig:qual-eval}
\end{figure}
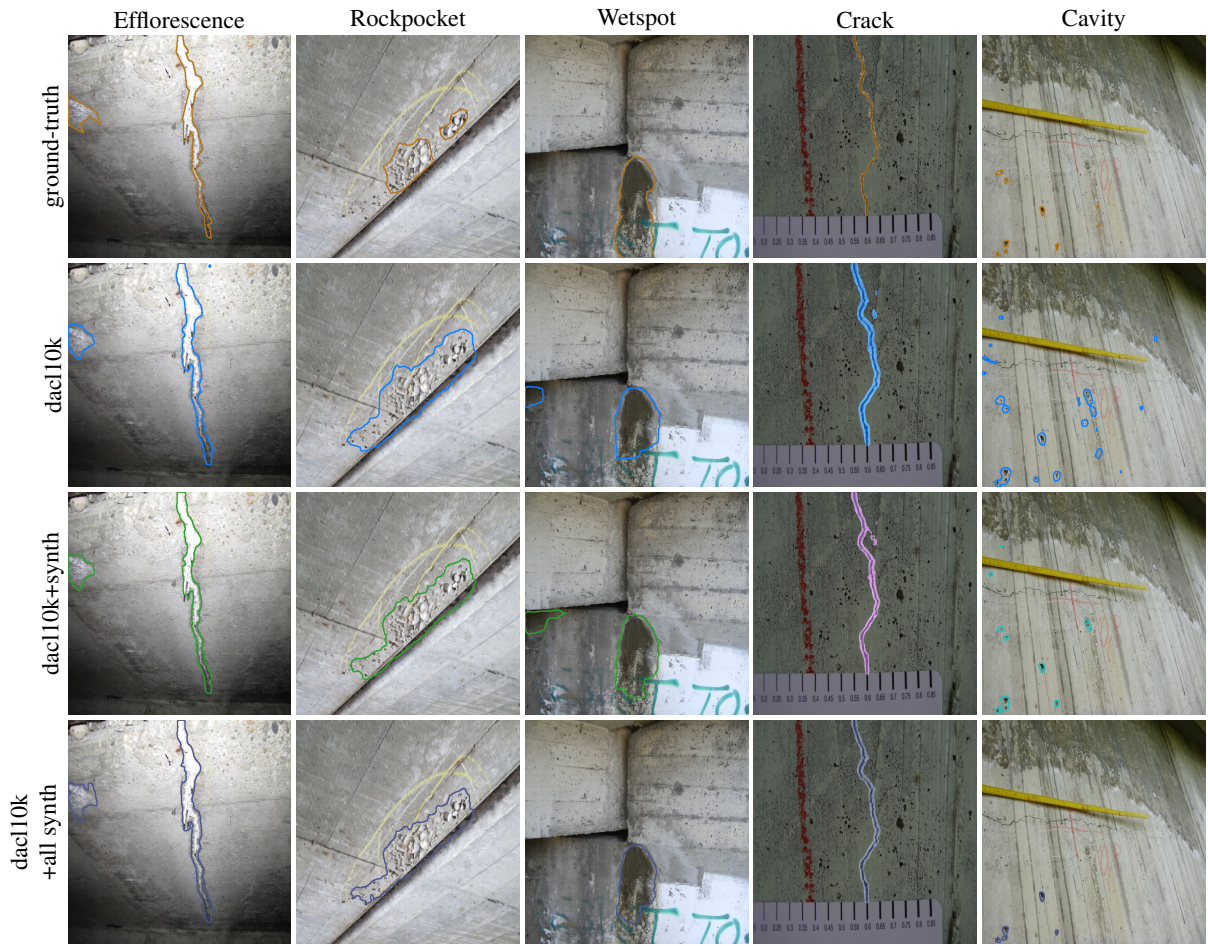

In some cases, the labels in the dacl10k dataset are overly coarse and not precise, which limits the effectiveness of quantitative evaluation metrics like IoU in reflecting the model's performance. Therefore, we complement our evaluation with qualitative results in Figure~\ref{fig:qual-eval}, providing a clearer assessment of the model's capability in handling the task.
The predictions in the second row originate from the baseline trained on ``dacl10k'' (blue). The predictions in the ``dacl10k+synth'' row originate from the model trained on dacl10k plus the synthetic split that specifically includes the according class. In this row the three most left columns show predictions by the model trained  on dacl10k+daclonsynth (green), followed by the models trained on dacl10k+synthcrack (pink) and dacl10k+synthcavity (turquoise) respectively. The bottom row displays predictions from the network that used all data (grey), or rather the most robust model according to Table~\ref{tab:testall_pert}.

The qualitative examples mostly underline the metrics reported in aforementioned Tables.
The predictions on \textit{Wetspot} by the most robust model show no false positives compared to the baseline dacl10k.  On \textit{Rockpocket} the prediction gets from top to bottom more accurate indicating that the additional synthetic data raises Precision. 
The \textit{Crack} prediction proves that the segmentation gets narrower, thus closer to the crack edges but still leaves room for improvement. Regarding the \textit{Cavity} sample, we observe that only the relevant cavities are marked, indicating higher accuracy.

%%%%%%%%% DISCUSSION
\section{Discussion}
\label{sec:limits}

%%% General
According to Table~\ref{tab:iou}, introducing synthetic data specific to certain defects, such as \textit{Cracks} (synthcrack) or \textit{Cavities} (synthcavity), in isolation, the model performs better than without additional extensions. However, when both synthetic \textit{Crack} and \textit{Cavity} data are combined, it results in the lowest scores with respect to accuracy and robustness if trained without the balancing from daclonsynth (see Table~\ref{tab:iou} and \ref{tab:testall_pert}). This suggests that while synthetic data can effectively address individual class imbalances, combining certain synthetic datasets can create conflicts that negatively impact model performance by introducing imbalances again. 
This is illustrated in Figure~\ref{fig:daclonsynth-barchart} and \ref{fig:counts-added}. \Eg, the number of pixels and shapes showing \textit{Efflorescence} in dacl10k is 40 million pixels and 3,350 polygons, while synthcrack introduces 147 million pixels and 1.7 million polygons showing \textit{Weathering}.  

%%% Weathering: 
The best performance on \textit{Weathering} in Table~\ref{tab:iou} is reported for the model trained exclusively on dacl10k, suggesting that synthetic \textit{Weathering} features may not fully capture the characteristics of its real-world counterpart, which strongly rely on the subjacent concrete. Style transfer techniques may help to bridge this domain gap by enhancing the realism of synthetic textures \cite{luan2017deepphotostyletransfer}.

%%%% synthcrack
Although the model's predictions on finecrack masks appear promising (Figure~\ref{fig:qual-eval}), the achieved IoU of 13\% indicates substantial room for improvement in terms of practical applicability. A closer inspection of the predictions by the model trained only on synthcrack (see Supplementary Material) reveals that crack segments are often disrupted when objects such as shadows, Wetspots, or Efflorescence are located adjacent to the crack edges. To improve realism and robustness, future work should focus on augmenting the synthetic pipeline with additional examples of such crack-bordering artifacts.

%%% synthcavity 
The isolated use of synthcavity proves insufficient for real-world Cavity detection, as shown in Table~\ref{tab:iou-finecr}. Nonetheless, its combination with real-world data improves performance (Table~\ref{tab:iou}), supporting its role as a supplementary rather than standalone training asset. Qualitative analysis of predictions made by the model trained only on synthcavity (see Supplementary Material) reveals common false positives arising, like in the case of synthcrack, from visually alien elements such as bolts, soil patches, or drainage components. Moreover, while the model successfully identifies small cavities, it consistently misses large ones -- especially those with sharp, irregular geometries. This limitation stems from the underrepresentation such \textit{Cavity} features in the synthcavity dataset. 
In order to generate more realistic synthetic \textit{Cavity} images, future work should focus on incorporating methods such as texture-sensitive preprocessing \cite{hess2021proceduralworldgenerationframework}, style transfer methods \cite{park2020contrastivelearningunpairedimagetoimage,prabhu2023bridgingsim2realgapcare} or domain randomization strategies \cite{tobin2017domainrandomizationtransferringdeep}.

\section{Conclusion}

In this work, we introduce \textit{synth-dacl}, a new set of three synthetic dataset extensions designed to support automated damage recognition in real-world bridge inspections. Alongside these extensions, we provide a cleaned version of the dacl10k dataset and introduce fine-resolution crack masks for the test set, enabling more precise evaluation. Through a series of experiments, we analyze the impact of synthetic data on segmentation performance and model robustness by combining the original dacl10k dataset with our synthetic extensions.

Our findings show that synthetic data can significantly improve performance, though its effectiveness depends on how it is integrated. Training exclusively on the \textit{synthcrack} dataset yields the best performance on fine-resolution crack prediction, as demonstrated in Table~\ref{tab:iou-finecr}. We also found that the \textit{synthcavity} extension improves accuracy only when combined with real-world data, particularly in the dacl10k+\textit{synthcavity} setup. Using all three synthetic extensions (\textit{daclonsynth}, \textit{synthcrack}, and \textit{synthcavity}) leads to the most robust models, especially under challenging conditions.

Beyond these empirical results, our study reveals an important insight: although synthetic data is a powerful tool for augmenting training sets and improving model performance, it is not always ready for seamless integration into real-world applications. Further research is necessary to develop methods for generating synthetic data that more accurately capture the visual complexity, texture diversity, and environmental variability present in real-world bridge inspection scenarios. Our work lays a foundation for this future exploration of synthetic data in structural inspection tasks and the development of robust damage recognition systems.

\section*{Acknowledgments}
We thank the institute for Distributed Intelligent Systems at the University of the Bundeswehr Munich for providing computing time at MonacumOne Cluster.

%%%%%%%%% SupplMat
\clearpage
\setcounter{section}{0}

\begin{center}
    {\LARGE\bfseries Supplementary Material}
\end{center}

\section{Data Availablility}
\label{sec:DataAvailability}
We make all data used in the underlying research openly accessible:
\begin{itemize}
    \item The updated version of dacl10k-v3~\cite{RQUOYN_2025}.
    \item The synthetic dataset extensions~\cite{9D6E4M_2025}.
\end{itemize}

\section{Implementation Details}
\label{sec:ImplementationDetails}

We utilize a Feature Pyramid Network (FPN)~\cite{kirillov2019panoptic} architecture combined with an ImageNet pretrained MaxViT-base backbone~\cite{tu2022maxvit}. Training and testing on the dacl10k dataset are performed at a resolution of $512\times512$. To ensure consistency with previous work~\cite{dacl10k}, we employ a weighted loss function that optimizes two objectives: semantic segmentation and classification loss. 
The mask loss is a combination of Binary Cross-Entropy (BCE) and Dice Loss. In addition, for the classification task, we use Binary Cross-Entropy to assure that the model does not ignore smaller classes present in the images. This is critical for dealing with imbalances where small classes might otherwise not be predicted. The auxiliary loss is added with a weight of $0.1$.
During training, we set the learning rate to $1\mathrm{e}{-5}$ and use a cosine learning rate scheduler to gradually reduce the learning rate over 30 epochs. To improve the robustness of the model, we apply basic augmentations during training, specifically random rotations and horizontal/vertical flipping. These augmentations help the model to better generalize to different perspectives and orientations in the dataset.

\section{Datasets}
\label{sec:stats}
Additional statistics and details about the dacl10k-v3 dataset and the synth-dacl extensions are shown below. Table~\ref{tab:samples} shows the number of images present in dacl10k and the synthetic extensions. 
\begin{table}[h]
    \caption{Image count and count of images showing Weathering for dacl10k and the synth-dacl extensions.}
    \label{tab:samples}
    \centering
    \begin{tabular}{lcc}
    \toprule
                 & Images & Weathered \\
                 \midrule
         dacl10k & 6,935  & 3,449 \\
         daclonsynth & 5,000 & 2,500 \\
         synthcrack & 5,000 & 2,500 \\
         synthcavity & 5,000 & 2,500 \\
         \bottomrule
    \end{tabular}
\end{table}

\subsection{dacl10k-v3}

\begin{table*}[h]
\caption{Overall statistics of the dataset at original resolution regarding pixel count (\#pixels), shape/polygon count (\#shape), image count (\#images), average number of shapes per image (\#shape/image), number of pixels per shape (\#pixels/shape), number of pixels per image (\#pixels/image), share of shapes (\%shape) and share of pixels (\%pixels). Midrules separate the classes according to their group affiliation.}
\label{tab:v3-overview}
\centering
\tiny
\begin{tabular}{lrrrrrrrr} %{L{11mm}R{11mm}R{6mm}R{6mm}R{11mm}R{10mm}R{10mm}R{6mm}R{6mm}}
\toprule
Class         & \#pixels       & \#shape & \#images & \#shape/image & \#pixels/shape & \#pixels/image & \%shape & \%pixels \\
\midrule
Crack         & 120,670,614    & 4,446      & 2,464    & 1.80             & 27,141           & 48,973         & 4.10       & 0.30     \\
ACrack        & 497,199,653    & 523        & 466      & 1.12             & 950,669          & 1,066,952      & 0.48       & 1.23     \\
Efflorescence & 1,045,865,294  & 5,022      & 2,181    & 2.30             & 208,257          & 479,535        & 4.63       & 2.59     \\
Rockpocket    & 60,669,651     & 751        & 446      & 1.68             & 80,785           & 136,031        & 0.69       & 0.15     \\
WConccor      & 225,428,183    & 492        & 366      & 1.34             & 458,187          & 615,924        & 0.45       & 0.56     \\
Hollowareas   & 788,944,545    & 1,900      & 1,564    & 1.21             & 415,234          & 504,440        & 1.75       & 1.95     \\
Cavity        & 156,930,825    & 11,661     & 1,722    & 6.77             & 13,458           & 91,133         & 10.76      & 0.39     \\
Spalling      & 1,055,540,626  & 12,272     & 4,705    & 2.61             & 86,012           & 224,344        & 11.32      & 2.61     \\
Restformwork  & 59,095,976     & 1,282      & 1,064    & 1.20             & 46,097           & 55,541         & 1.18       & 0.15     \\
\midrule
Wetspot       & 536,569,266    & 1,974      & 1,343    & 1.47             & 271,818          & 399,530        & 1.82       & 1.33     \\
Rust          & 824,807,051    & 17,679     & 4,883    & 3.62             & 46,655           & 168,914        & 16.31      & 2.04     \\
Graffiti      & 470,487,206    & 2,705      & 1,171    & 2.31             & 173,932          & 401,782        & 2.50       & 1.16     \\
Weathering    & 2,865,923,796  & 5,021      & 3,449    & 1.46             & 570,787          & 830,943        & 4.63       & 7.09     \\
\midrule
ExposedRebars & 64,287,097     & 2,490      & 1,109    & 2.25             & 25,818           & 57,969         & 2.30       & 0.16     \\
Bearing       & 638,064,849    & 1,506      & 1,038    & 1.45             & 423,682          & 614,706        & 1.39       & 1.58     \\
EJoint        & 422,636,952    & 602        & 538      & 1.12             & 702,055          & 785,571        & 0.56       & 1.05     \\
Drainage      & 466,415,500    & 2,016      & 1,470    & 1.37             & 231,357          & 317,289        & 1.86       & 1.15     \\
PEquipment    & 1,482,991,703  & 2,314      & 1,888    & 1.23             & 640,878          & 785,483        & 2.14       & 3.67     \\
JTape         & 64,621,780     & 1,269      & 1,092    & 1.16             & 50,923           & 59,177         & 1.17       & 0.16     \\
\midrule
Background    & 30,288,963,831 & 32,448     & 9,852    & 3.29             & 933,462          & 3,074,397      & 29.94      & 74.91    \\
\bottomrule
\end{tabular}
\end{table*}

The following paragraphs, especially the statistics on classes that are in the main focus of synth-dacl are analyzed, subdivided into the three synthetic extensions (see Table~\ref{tab:v3-overview}). 

\textbf{daclonsynth.} 
% Efflorescence + Spalling (dominant)
\textit{Efflorescence} and \textit{Spalling} emerge as dominant classes in terms of pixel coverage, each contributing over 1 billion pixels (2.59\% and 2.61\% of the dataset, respectively). This is complemented by a significant number of shapes (5,022 for \textit{Efflorescence} and 12,272 for \textit{Spalling}).
% Rust
Rust has the highest number of shapes in this group (17,679 shapes, 16.31\% of total shapes). However, the average size per shape (46,655 pixels) indicates that Rust's polygons are predominantly small.
% Wetspot and Hollowareas 
\textit{Wetspot} and \textit{Hollowarea} are moderately represented compared to other classes in synth-dacl, with pixel counts of 537M (1.33\%) and 789M (1.95\%). The average \textit{Wetspot} contains 271,818 pixels per shape. \textit{Hollowareas} are half the size with 415,234 pixels per shape.
% Rockpocket and ExposedRebars
\textit{Rockpocket} and \textit{ExposedRebars} are underrepresented, with ~60M pixels each (~0.15\% of total dataset pixels). Despite their lower prevalence, these classes maintain moderate shape complexity, with average pixels per shape of 80,785 and 25,818, respectively.

\textbf{Crack.} 
The class \textit{Crack} is characterized by a pixel count of 120.7M pixels, which is 0.30\% of the dataset, and a moderate number of shapes (4,446, 4.10\%). On average, \textit{Crack} polygons have a size of 27,141 pixels/shape.

\textbf{Cavity.} 
The \textit{Cavity} class has a high number of shapes (11,661, 10.76\% of all shapes), making it the most shape-dense class in synth-dacl. However, these shapes are remarkably small, with an average size of 13,458 pixels. 
% The average size of a Cavity polygon in the synthcavity extension is 19 pixels. 
Despite the modest total number of pixels (156.9M, 0.39\%), the high frequency of \textit{Cavity} shapes (6.77\% shapes/image) suggests that when \textit{Cavity} is present, the entire concrete surface is usually affected. Cavities are residual air voids on the concrete surface. They are usually a ``cosmetic'' defect and do not indicate any deficiencies in structural integrity. 

For dacl10k-v3 we use the same partitioning as for dacl10-v2, which is based on K-means clustering (20 clusters), with proportional sampling across training (70\%), validation (10\%), test-dev (10\%), and test-priv (10\%) splits.

\subsection{Synthetic Addons}
\subsubsection{daclonsynth}
\begin{figure}
    \centering
    \includegraphics[width=.8\linewidth]{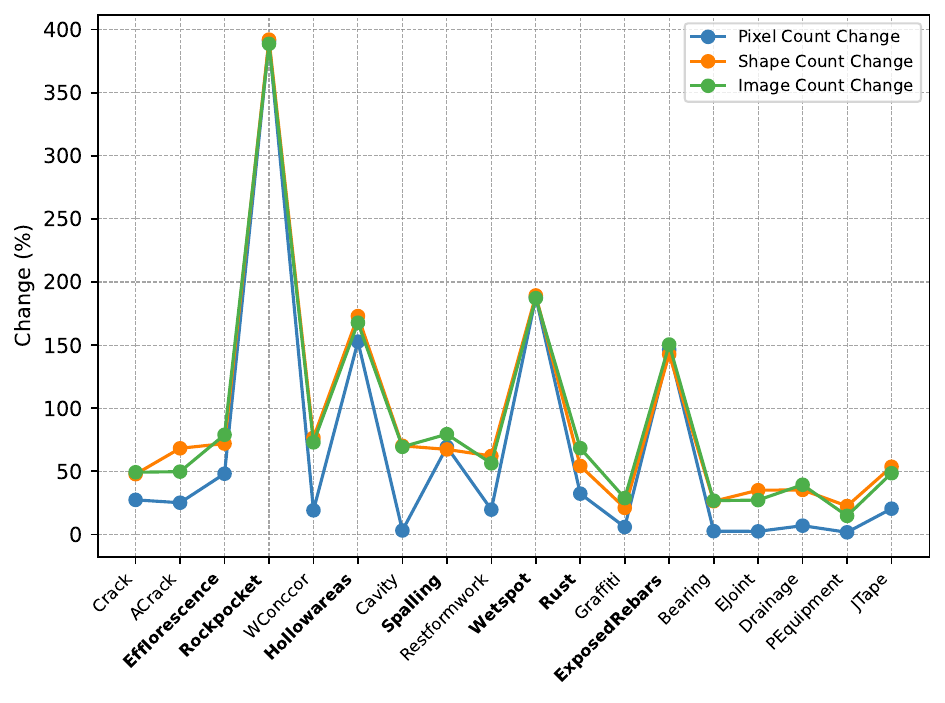}
    \caption{Distribution shift on class level regarding pixel, shape and image count. Classes considered in daclonsynth are in bold.}
    \label{fig:daclonsynth-change}
\end{figure}

% Classes that were not considered in daclonsynth and why 
A detailed discussion of the daclonsynth dataset follows in this section. Figure~\ref{fig:daclonsynth-change} displays the change of pixels, shapes, and images by adding dalonsynth to dacl10k, in percentage. 
The most significant changes occur in classes such as Exposed Rebars, Hollowarea, \textit{Wetspot}, and \textit{Rockpocket} with large increases in pixel, shape, and image counts. For example, the Rocketpocket class sees a striking jump, especially in image count change, reaching nearly 400\% (see Figure~\ref{fig:daclonsynth-change}).
Although they are not in the focus of daclonsynth, defects like \textit{Crack}, \textit{ACrack} or \textit{Cavity} are also added. This is due to the overlapping of defects as mentioned in the main article. For most of these ``unintentional'' defects, a gain in pixel and shape count of about 50\% can be observed. 

\begin{figure}
    \centering
    \includegraphics[width=0.8\linewidth]{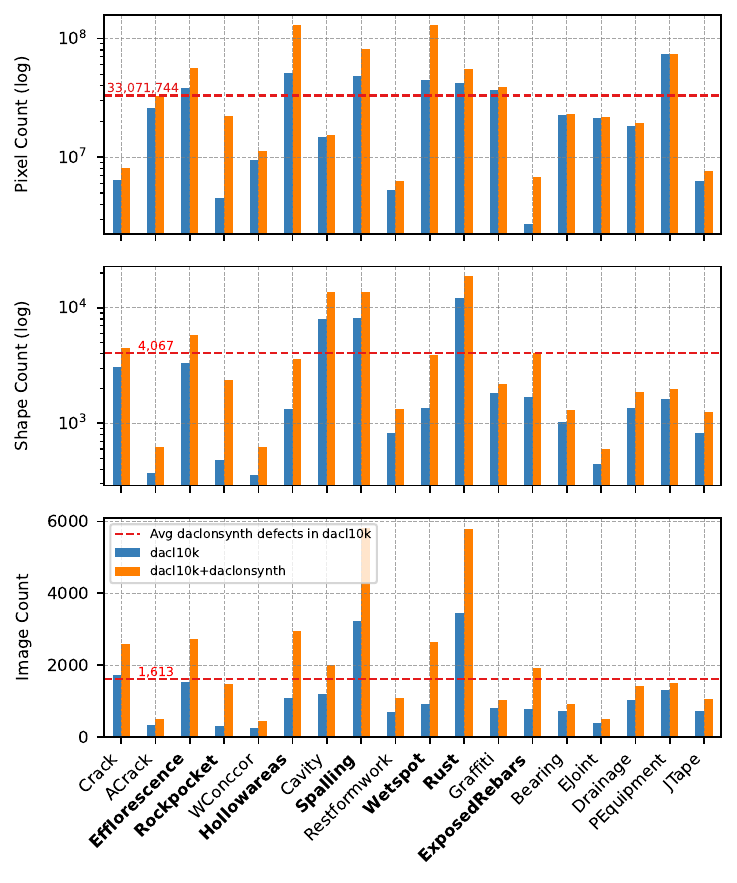}
    \caption{Comparison of classwise pixel count (top) and shape count (bottom) of dacl10k (blue) and the combination of dacl10k and daclonsynth (orange) dataset. The red dashed horizontal line marks the average pixel, or rather shape count, of daclonsynth classes (bold) which was aimed to reach through adding daclonsynth.}
    \label{fig:daclonsynth-big-barchart}
\end{figure}

Figure~\ref{fig:daclonsynth-big-barchart} presents the change in pixel, shape and image distribution at the class level. 

\textbf{Comment on daclonsynth class selection.} Concrete defects such as \textit{Crack}, \textit{ACrack} and \textit{Cavity} are not part of daclonsynth because they are considered in synthcrack and synthcavity. Annotations of Washouts/Concrete corrosion (WConccor) are ambiguous and need to be cleaned by a domain expert and are therefore neglected in this work. Restformwork would have required synthesized joints between concrete components and the exact fitting of polystyrene panels into the corresponding joint. Due to this complexity, this class was not taken into account.
Furthermore, the bridge components Bearing, \textit{Expansion Joint} (EJoint), Drainage, \textit{Protective Equipment} (PEquipment), JTape (JTape) are not in the scope of daclonsynth, too. Objects of this group do not appear on concrete. For instance, \textit{Protective Equipment}, such as railings on the side of the bridge, can be present in front of sky, forest, or water. Similarly, components like \textit{Expansion Joint} and joint tape are usually photographed from a top-down perspective, situated between asphalt surfaces on bridge decks. Moreover, these classes were excluded because the model already performs exceptionally well on them compared to other categories.

\section{Perturbations}
\label{sec:perts}
\begin{figure*}
    \centering
    \includegraphics[width=1\linewidth]{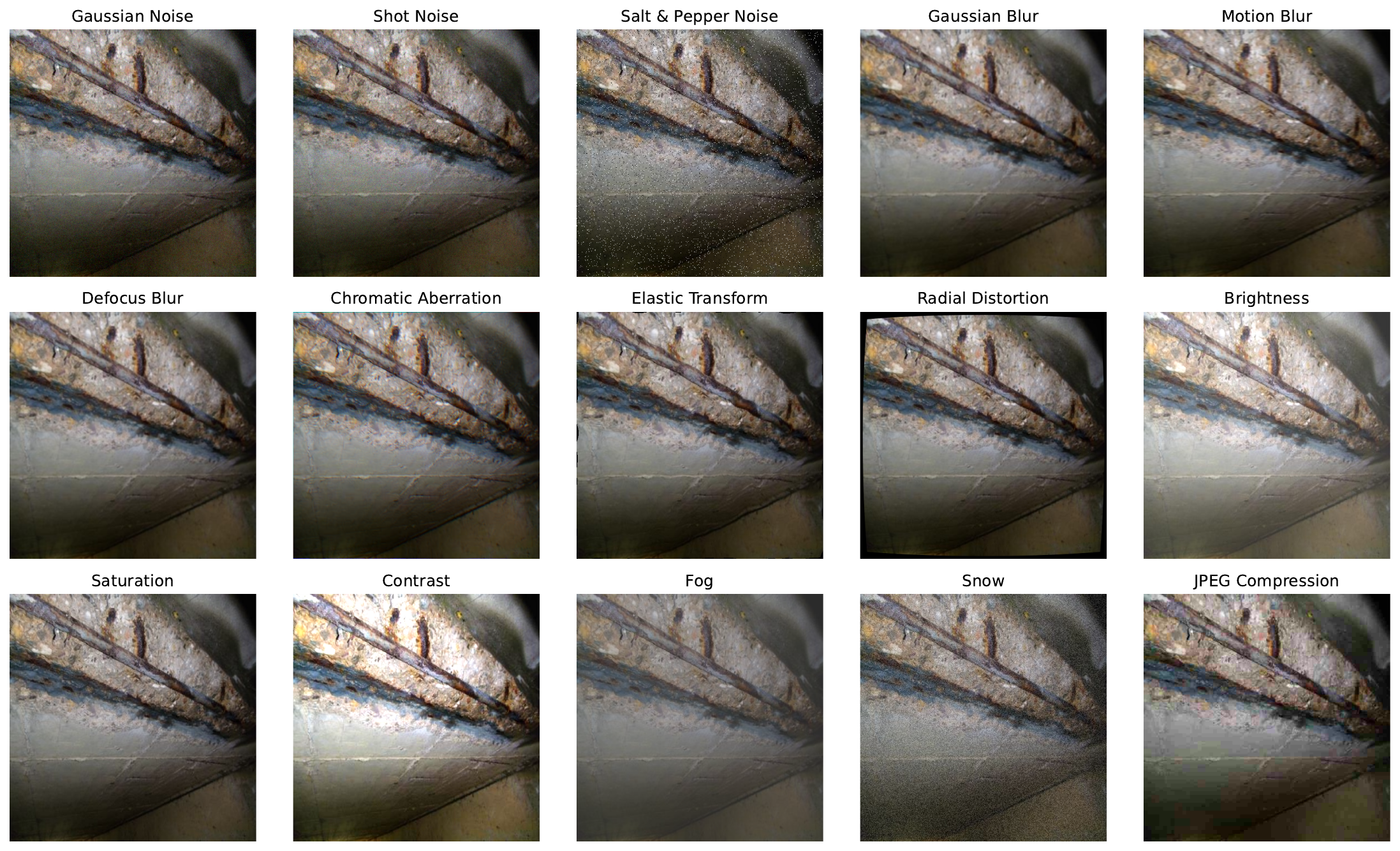}
    \caption{Illustration of the fifteen perturbation types utilized in experiments chapter, following \cite{WANG2024110685}.}
    \label{fig:perturb-overview}
\end{figure*}

\begin{figure*}
    \centering
    \includegraphics[width=.9\linewidth]{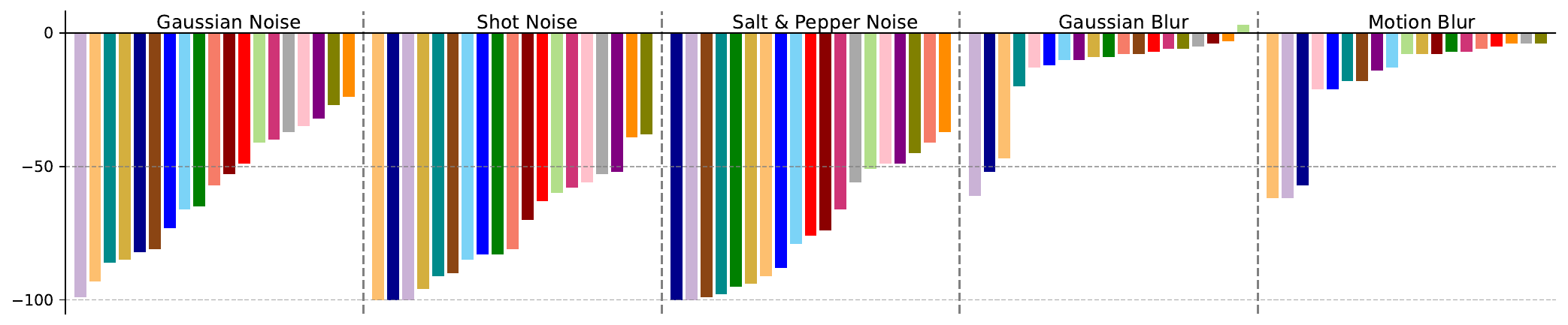}
    \includegraphics[width=.9\linewidth]{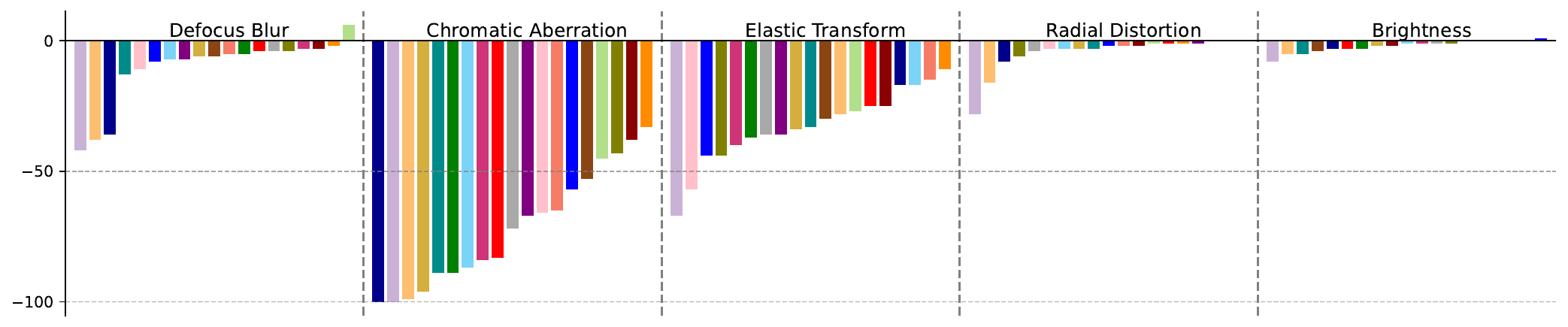}
    \includegraphics[width=.9\linewidth]{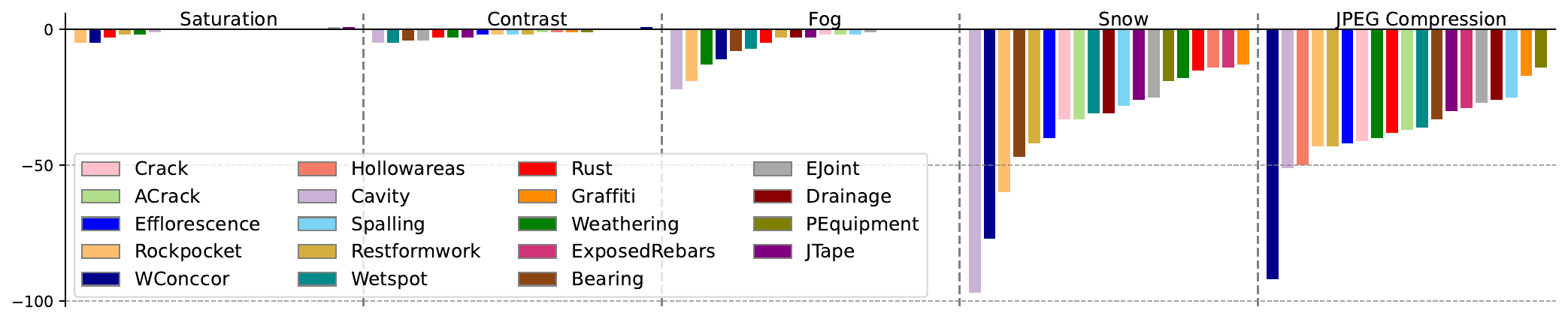}
    \caption{IoU drop in percent of the most robust model, trained on dacl10k and all synthetic extensions,
    % (see Table~\ref{tab:testall_pert}) 
    for each perturbation and class in the dacl10k test set.}
    \label{fig:iou_pert}
\end{figure*}

In Figure~\ref{fig:perturb-overview} all 15 perturbations are demonstrated using one example drawn from dacl10k. The sample displays \textit{Spalling}, \textit{Exposed Reinforcement Bars} (ExposedRebars) that are corroded (\textit{Rust}). 
Figure~\ref{fig:iou_pert} shows the IoU drop in percent of the ``most robust'' model that achieved the best overall performance on the perturbed data. It can be observed that among all classes, the concrete defects \textit{Cavity}, \textit{Rockpocket}, and \textit{WConccorr} are the most affected, compared to others that are usually sensitive to only one or two image defects.
The most resistant class is Graffiti and the building components such as \textit{Protective Equipment} (PEquipment), \textit{Drainage} or \textit{Expansion Joint} (EJoint). In terms of shape and texture, the components are more unique, as opposed to concrete defects, which often have a very similar visual appearance. 
Regarding \textit{Crack}, the Elastic Transform significantly disturbs the model, likely due to the distortion of the distinct zigzag pattern and sharp corners characteristic of \textit{Cracks}.

In summary, the listed perturbations strongly affect the model performance (some classes by 100\%), compared to other models, \eg, SAM, as analyzed in Wang~\etal \cite{WANG2024110685}, who report a maximum IoU drop of about 30\% on the Road Extraction dataset \cite{Demir2018DeepGlobe2A}.

\section{Predictions Domain-Partitioned Evaluation}
\begin{figure*}[!ht]
\centering
\resizebox{\linewidth}{!}{
\begin{tikzpicture}
% Set image width and spacing
\def\imgwidth{3.5cm}  % Image width
\def\xstart{2}  % Starting x position
\def\ystart{8}  % Starting y position
\def\xoffset{3.6}  % Horizontal offset between images
\def\yoffset{-3.6}  % Vertical offset between rows
% Adjusted y-offset for labels
\def\labeloffset{2}  % Additional y-offset for labels to prevent overlap
% First row (Sample)
\node at (\xstart-2, \ystart) {\rotatebox{90}{\small test sample}};
\node at (\xstart, \ystart) {\includegraphics[width=\imgwidth]{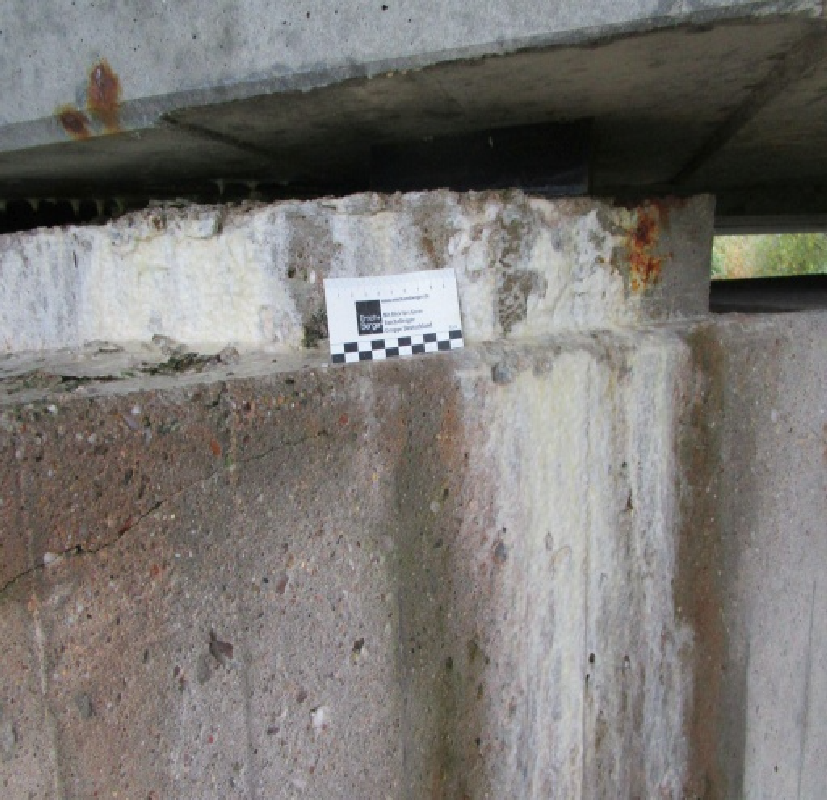}};
\node at (\xstart+\xoffset, \ystart) {\includegraphics[width=\imgwidth]{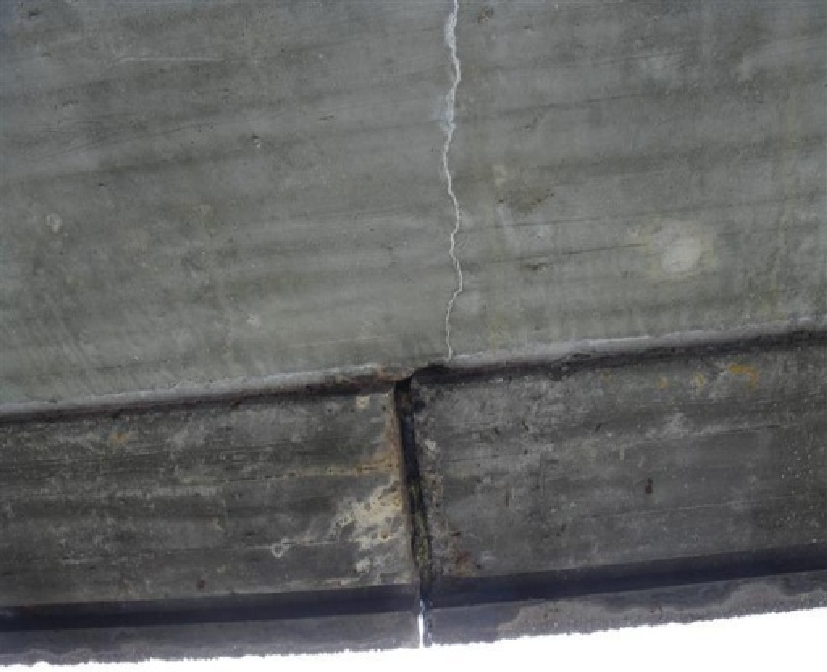}};
\node at (\xstart+\xoffset*2, \ystart) {\includegraphics[width=\imgwidth]{img/Figure1/00046.jpg}};
\node at (\xstart+\xoffset*3, \ystart) {\includegraphics[width=\imgwidth]{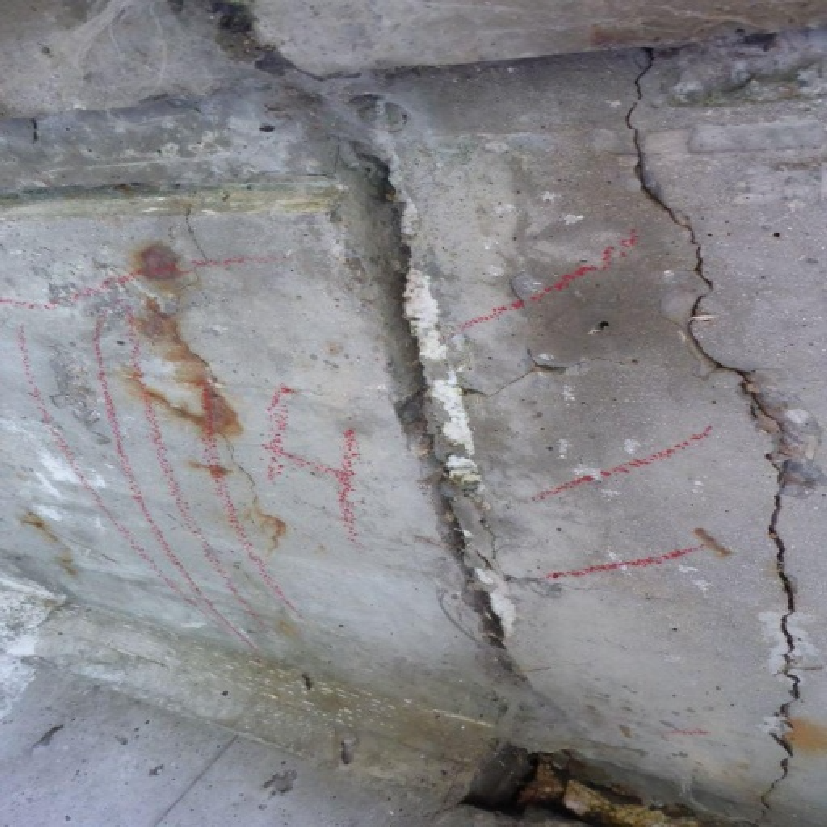}};

% Second row (ground-trouth)
\node at (\xstart-2, \ystart+\yoffset) {\rotatebox{90}{\small ground-truth}};
\node at (\xstart, \ystart+\yoffset) {\includegraphics[width=\imgwidth]{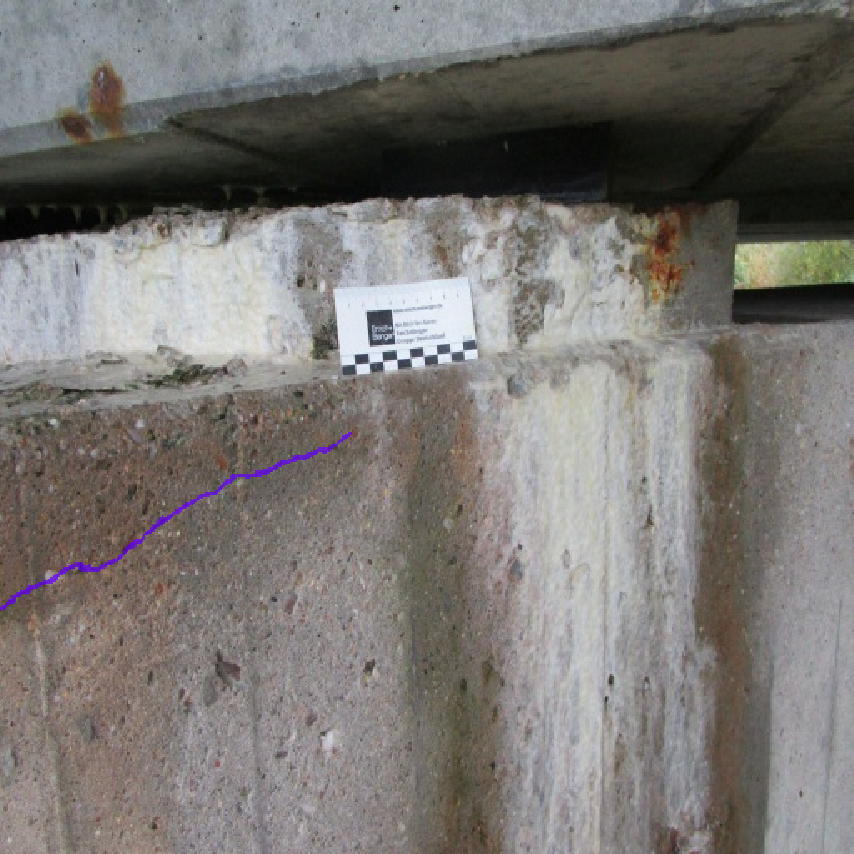}};
\node at (\xstart+\xoffset, \ystart+\yoffset) {\includegraphics[width=\imgwidth]{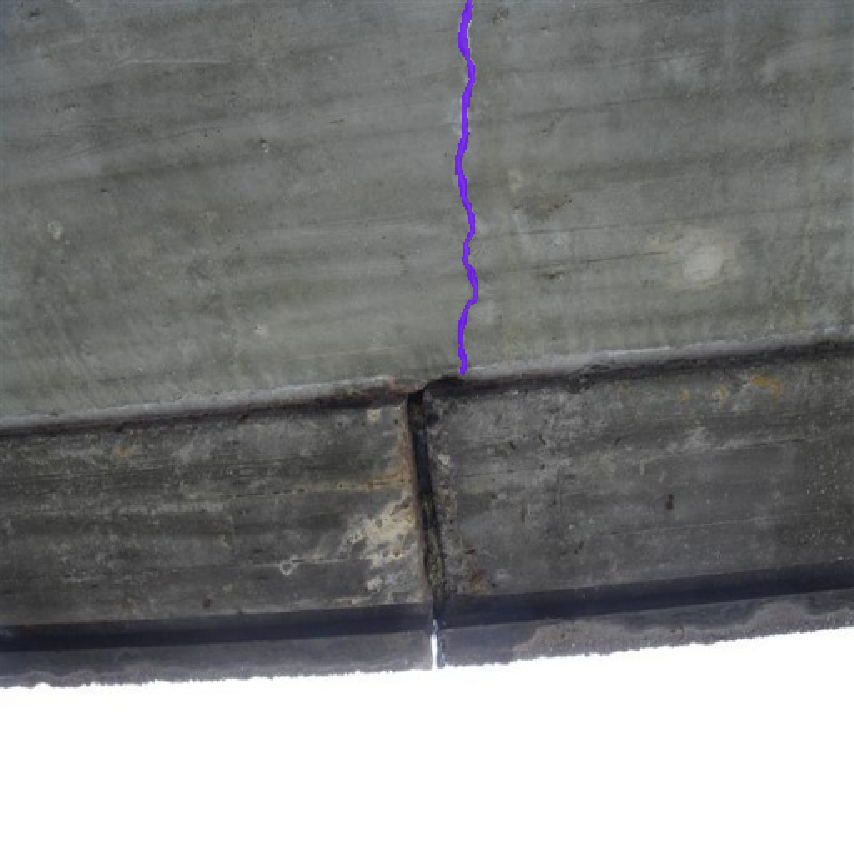}};
\node at (\xstart+\xoffset*2, \ystart+\yoffset) {\includegraphics[width=\imgwidth]{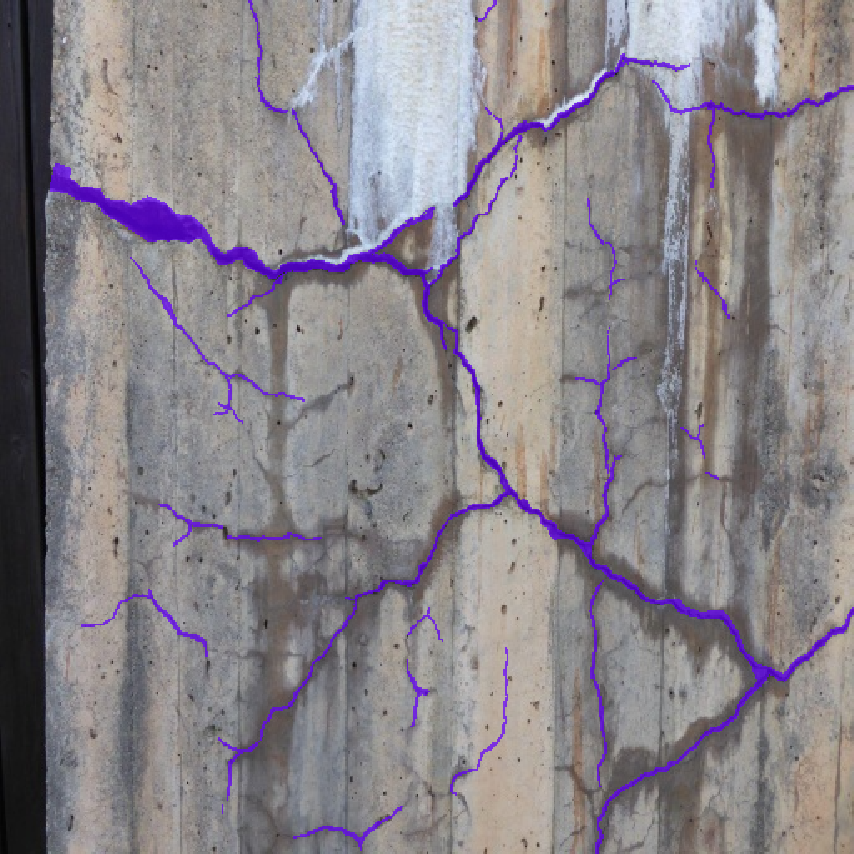}};
\node at (\xstart+\xoffset*3, \ystart+\yoffset) {\includegraphics[width=\imgwidth]{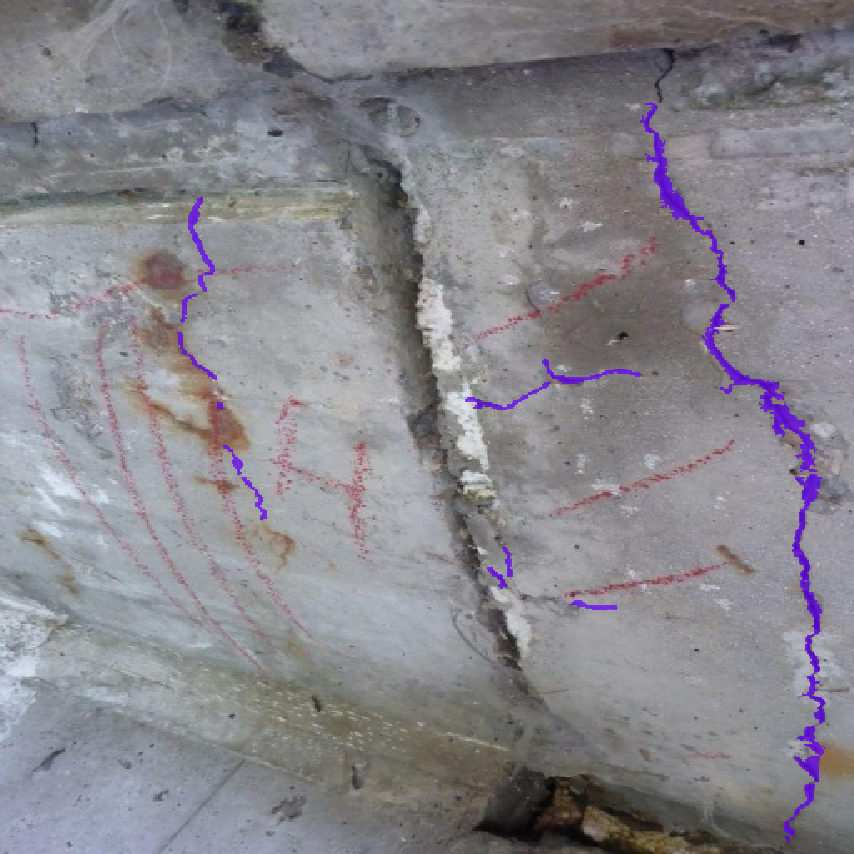}};

% Third row (ground-trouth)
\node at (\xstart-2, \ystart+\yoffset*2) {\rotatebox{90}{\small prediction}};
\node at (\xstart, \ystart+\yoffset*2) {\includegraphics[width=\imgwidth]{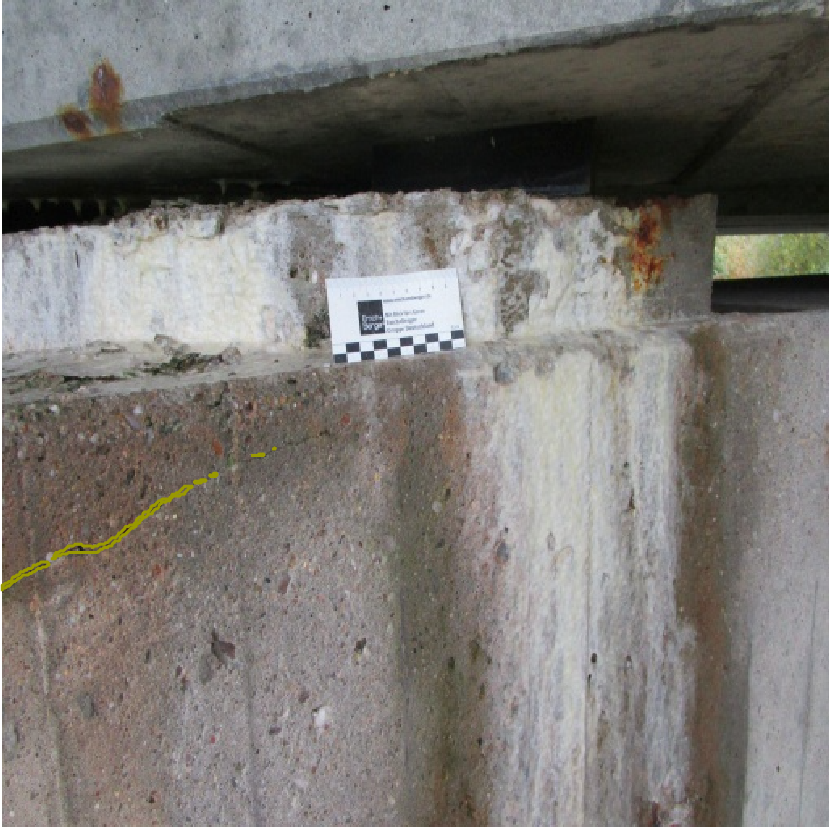}};
\node at (\xstart+\xoffset, \ystart+\yoffset*2) {\includegraphics[width=\imgwidth]{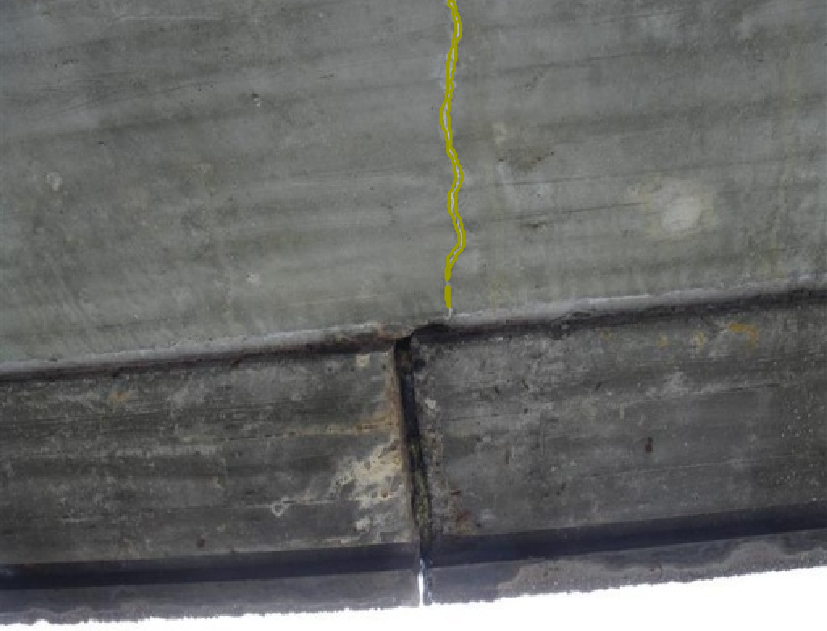}};
\node at (\xstart+\xoffset*2, \ystart+\yoffset*2) {\includegraphics[width=\imgwidth]{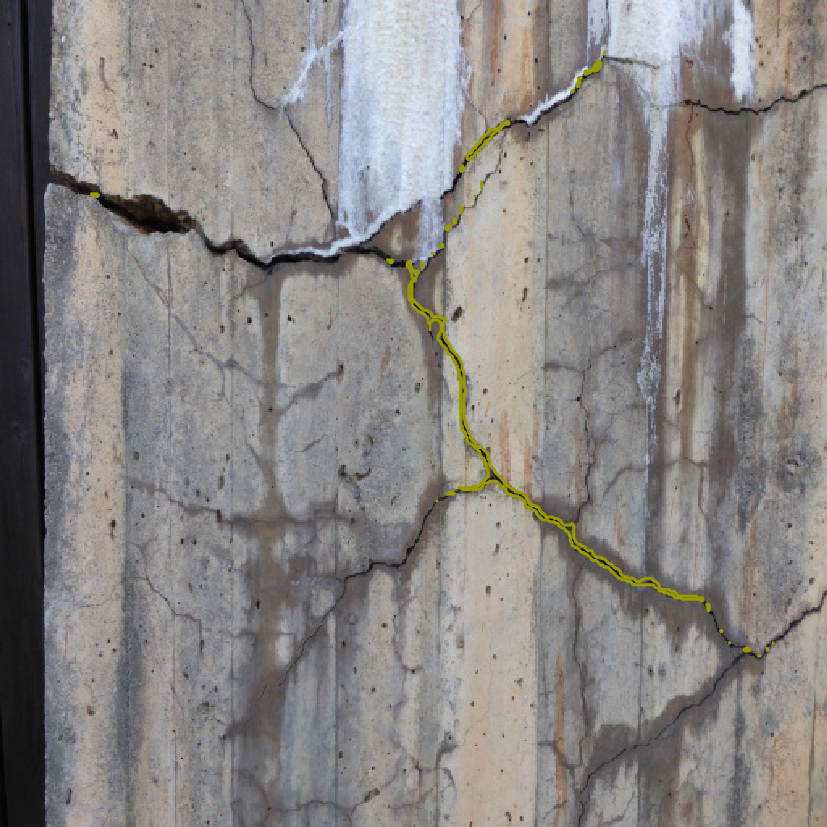}};
\node at (\xstart+\xoffset*3, \ystart+\yoffset*2) {\includegraphics[width=\imgwidth]{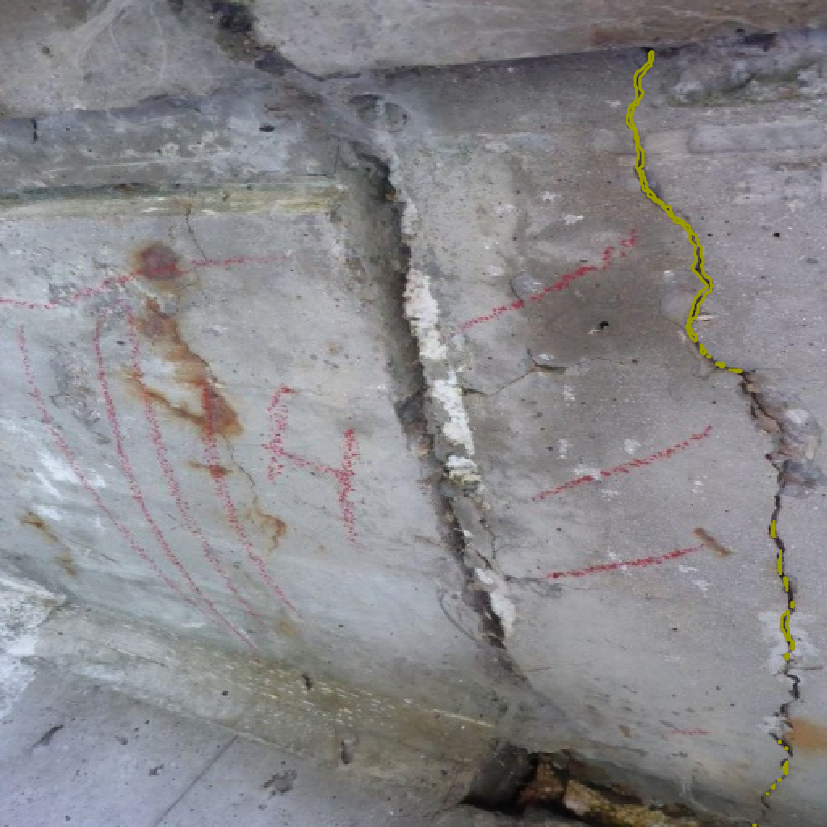}};
% Labels for the columns (top row)
\node at (\xstart, \ystart+\labeloffset) {\small \greencheck};
\node at (\xstart+\xoffset, \ystart+\labeloffset) {\small \greencheck};
\node at (\xstart+\xoffset*2, \ystart+\labeloffset) {\small \yellowcross};
\node at (\xstart+\xoffset*3, \ystart+\labeloffset) {\small \yellowcross}; % new label

\end{tikzpicture}
}

\caption{Rated (column header) predictions of the model trained on synthcrack only.}
\label{fig:pred-synthcrack}
\end{figure*}

\begin{figure*}[!ht]
\centering
\resizebox{\linewidth}{!}{
\begin{tikzpicture}
% Set image width and spacing
\def\imgwidth{3.5cm}  % Image width
\def\xstart{2}  % Starting x position
\def\ystart{8}  % Starting y position
\def\xoffset{3.6}  % Horizontal offset between images
\def\yoffset{-3.6}  % Vertical offset between rows
% Adjusted y-offset for labels
\def\labeloffset{2}  % Additional y-offset for labels to prevent overlap
% First row (Sample)
\node at (\xstart-2, \ystart) {\rotatebox{90}{\small test sample}};
\node at (\xstart, \ystart) {\includegraphics[width=\imgwidth]{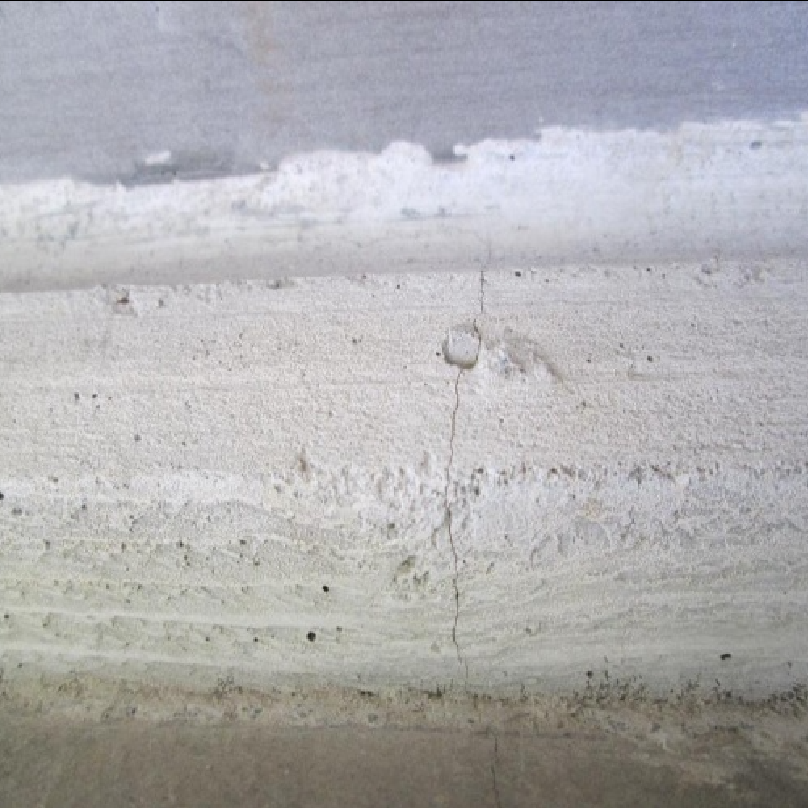}};
\node at (\xstart+\xoffset, \ystart) {\includegraphics[width=\imgwidth]{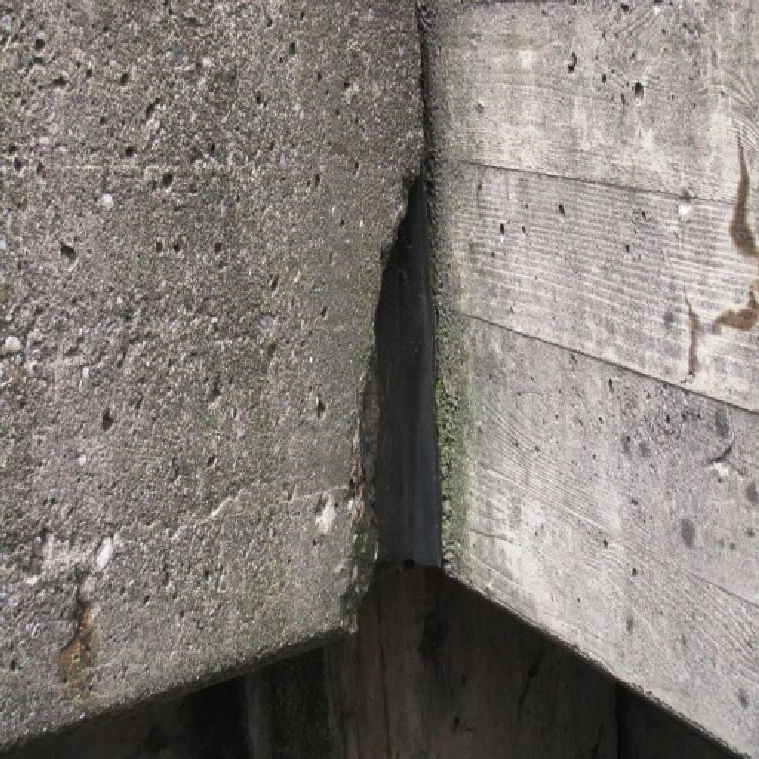}};
\node at (\xstart+\xoffset*2, \ystart) {\includegraphics[width=\imgwidth]{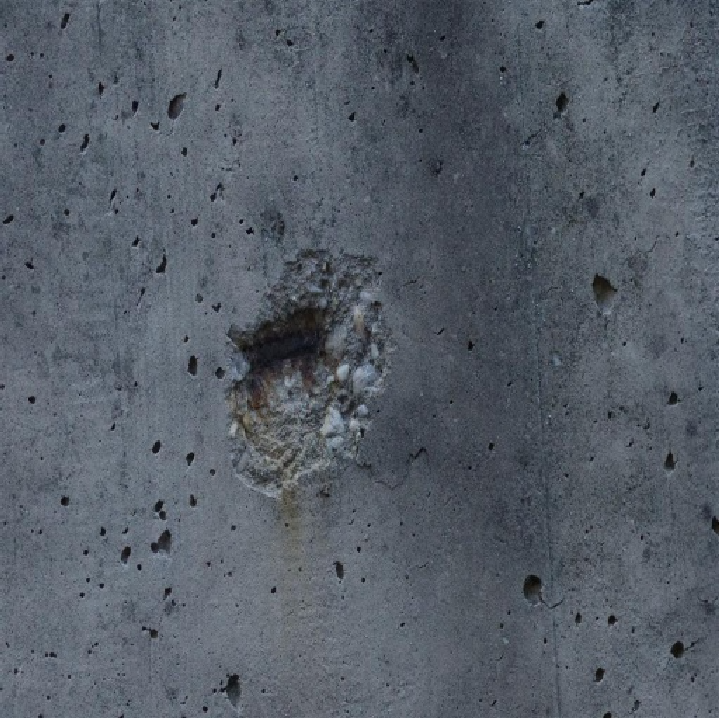}};
\node at (\xstart+\xoffset*3, \ystart) {\includegraphics[width=\imgwidth]{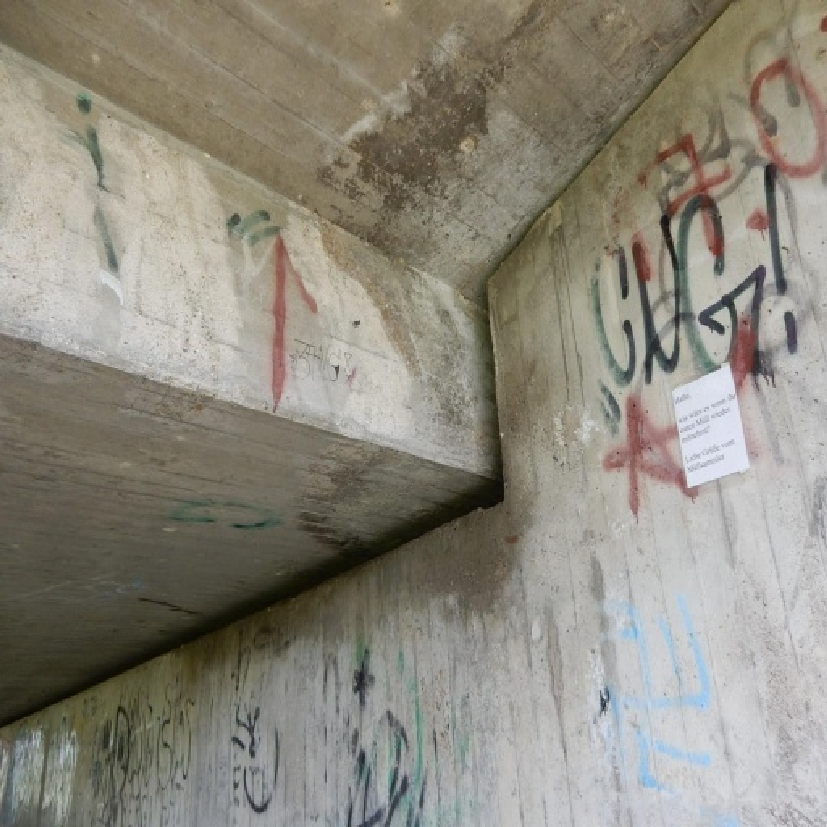}};
% Second row (ground-trouth)
\node at (\xstart-2, \ystart+\yoffset) {\rotatebox{90}{\small prediction+ground-truth}};
\node at (\xstart, \ystart+\yoffset) {\includegraphics[width=\imgwidth]{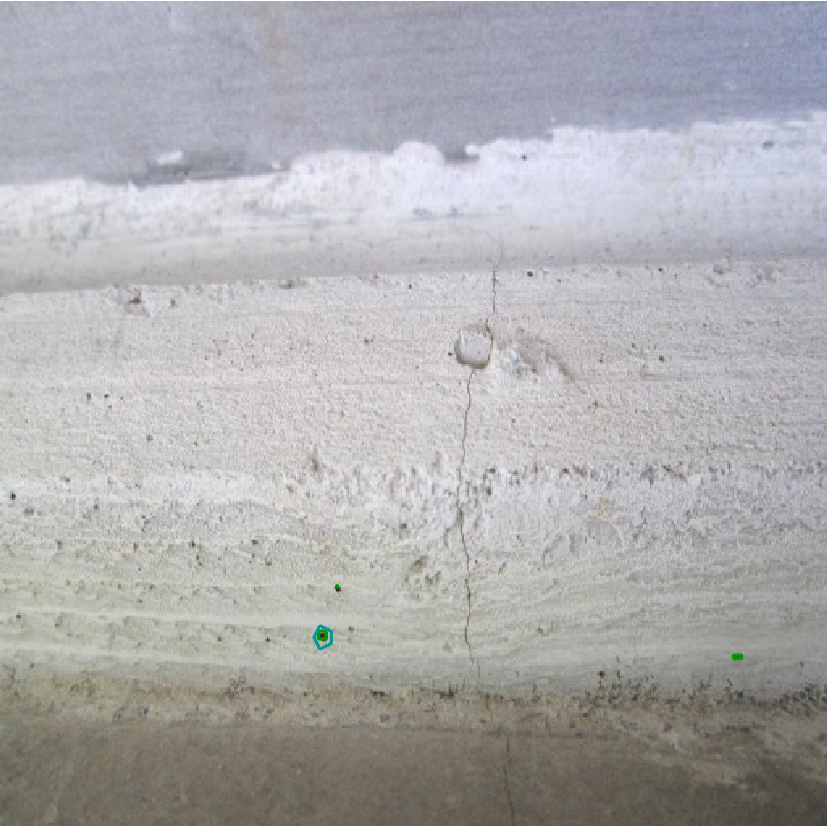}};
\node at (\xstart+\xoffset, \ystart+\yoffset) {\includegraphics[width=\imgwidth]{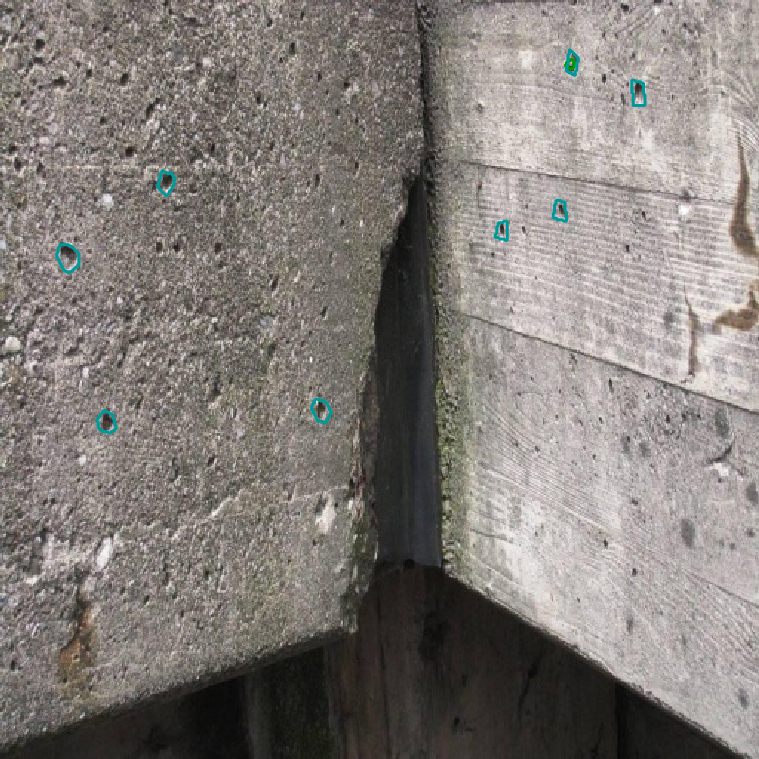}};
\node at (\xstart+\xoffset*2, \ystart+\yoffset) {\includegraphics[width=\imgwidth]{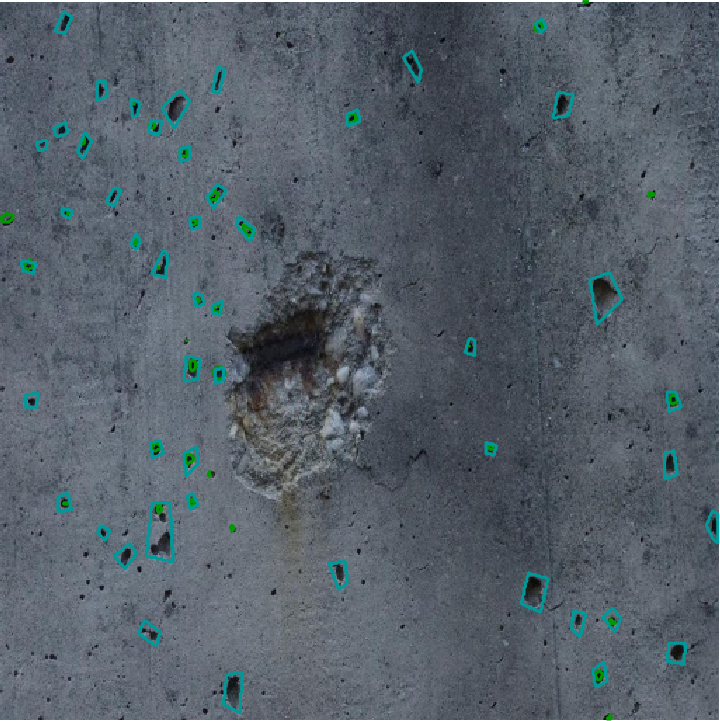}};
\node at (\xstart+\xoffset*3, \ystart+\yoffset) {\includegraphics[width=\imgwidth]{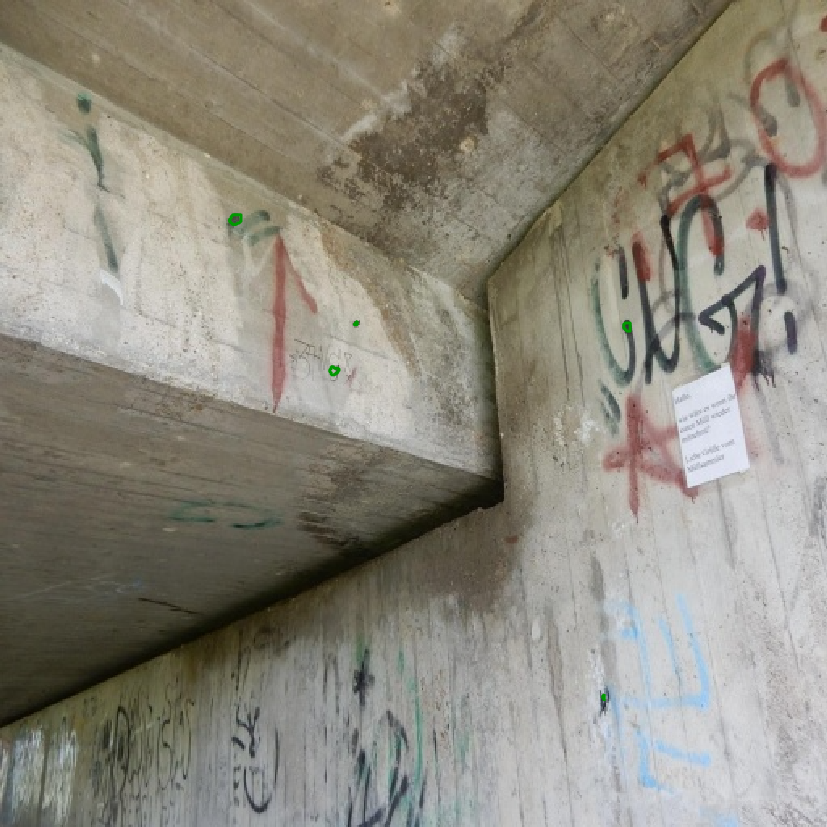}};
% Labels for the columns (top row)
\node at (\xstart, \ystart+\labeloffset) {\small \greencheck};
\node at (\xstart+\xoffset, \ystart+\labeloffset) {\small \yellowcross};
\node at (\xstart+\xoffset*2, \ystart+\labeloffset) {\small \yellowcross};
\node at (\xstart+\xoffset*3, \ystart+\labeloffset) {\small \redcross}; % new label

\end{tikzpicture}
}

\caption{Rated predictions (column header) of the model trained on synthcavity only. Predictions are green and ground-truth is blue.}
\label{fig:pred-synthcavity}
\end{figure*}
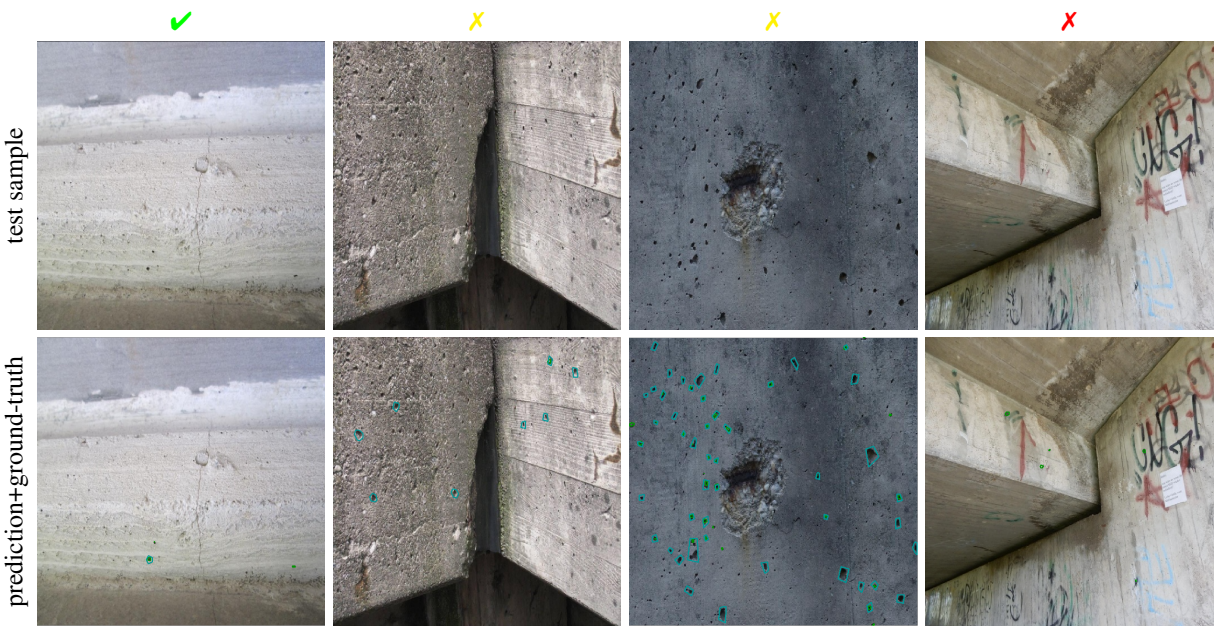

Figure~\ref{fig:pred-synthcavity} and Figure~\ref{fig:pred-synthcrack} present the predictions and ground-truth annotations for models trained exclusively on their respective synthetic extensions. The predictions produced by the synthcrack model demonstrate good performance on cracks that are relatively straight and unbranched. However, the model's accuracy decreases when cracks exhibit complex branching or when wet spots or efflorescence intersect the crack edges, often leading to incomplete or interrupted predictions.

For the synthcavity model, fine-grained and small cavities are partially predicted. In contrast, large cavities with sharp boundaries are often not reliably detected, indicating limitations in generalizing to more geometrically distinct damage patterns.  

%Bibliography
\bibliographystyle{unsrt}  
\bibliography{refs}

\begin{thebibliography}{10}

\bibitem{dacl10k}
Johannes Flotzinger, Philipp~J. R\"osch, and Thomas Braml.
\newblock dacl10k: Benchmark for semantic bridge damage segmentation.
\newblock In {\em Proceedings of the IEEE/CVF Winter Conference on Applications of Computer Vision (WACV)}, pages 8626--8635, January 2024.

\bibitem{artbridge2024}
{American Road \& Transportation Builders Association (ARTBA)}.
\newblock {ARTBA} bridge report, 2024.
\newblock Accessed: 2024-10-17.

\bibitem{brueckstat}
Federal Highway Research~Institute (BaSt).
\newblock Brückenstatistik.
\newblock \url{https://www.bast.de/DE/Statistik/Bruecken/Brueckenstatistik.pdf}, 5 2025.
\newblock [Accessed 15-05-2025].

\bibitem{Phares2004}
Brent~M. Phares, Glenn~A. Washer, Dennis~D. Rolander, Benjamin~A. Graybeal, and Mark Moore.
\newblock Routine highway bridge inspection condition documentation accuracy and reliability.
\newblock {\em Journal of Bridge Engineering}, 9(4):403--413, July 2004.

\bibitem{LIU2023100167}
Pengkun Liu, Ying Shi, Ruoxin Xiong, and Pingbo Tang.
\newblock Quantifying the reliability of defects located by bridge inspectors through human observation behavioral analysis.
\newblock {\em Developments in the Built Environment}, 14:100167, 2023.

\bibitem{abdallah2021bridge}
Abdelrahman~M. Abdallah.
\newblock {\em A Study on Bridge Inspections: Identifying Barriers to New Practices and Providing Strategies for Change}.
\newblock Phd dissertation, Colorado State University, 2021.

\bibitem{benz2022image}
Christian Benz and Volker Rodehorst.
\newblock Image-based detection of structural defects using hierarchical multi-scale attention.
\newblock In {\em DAGM German Conference on Pattern Recognition (GCPR)}, pages 337--353. Springer, 2022.

\bibitem{HE2024102586}
Zengsheng He, Cheng Su, and Yichuan Deng.
\newblock A novel mo-yolov4 for segmentation of multi-class bridge damages.
\newblock {\em Advanced Engineering Informatics}, 62:102586, 2024.

\bibitem{dacl_challenge}
Johannes Flotzinger, Philipp~J. R\"osch, Christian Benz, Muneer Ahmad, Murat Cankaya, Helmut Mayer, Volker Rodehorst, Norbert Oswald, and Thomas Braml.
\newblock Dacl-challenge: Semantic segmentation during visual bridge inspections.
\newblock In {\em Proceedings of the IEEE/CVF Winter Conference on Applications of Computer Vision (WACV) Workshops}, pages 716--725, January 2024.

\bibitem{Benz_2024_CVPR}
Christian Benz and Volker Rodehorst.
\newblock Omnicrack30k: A benchmark for crack segmentation and the reasonable effectiveness of transfer learning.
\newblock In {\em Proceedings of the IEEE/CVF Conference on Computer Vision and Pattern Recognition (CVPR) Workshops}, pages 3876--3886, June 2024.

\bibitem{kulkarni2022crackseg9k}
Shreyas Kulkarni, Shreyas Singh, Dhananjay Balakrishnan, Siddharth Sharma, Saipraneeth Devunuri, and Sai Chowdeswara~Rao Korlapati.
\newblock Crackseg9k: a collection and benchmark for crack segmentation datasets and frameworks.
\newblock In {\em European Conference on Computer Vision}, pages 179--195. Springer, 2022.

\bibitem{poucin2021boosting}
Florentin Poucin, Andrea Kraus, and Martin Simon.
\newblock Boosting instance segmentation with synthetic data: A study to overcome the limits of real world data sets.
\newblock In {\em Proceedings of the IEEE/CVF International Conference on Computer Vision}, pages 945--953, 2021.

\bibitem{dwibedi2017cut}
Debidatta Dwibedi, Ishan Misra, and Martial Hebert.
\newblock Cut, paste and learn: Surprisingly easy synthesis for instance detection.
\newblock In {\em Proceedings of the IEEE international conference on computer vision}, pages 1301--1310, 2017.

\bibitem{al2023usability}
Yasmina Al~Khalil, Sina Amirrajab, Cristian Lorenz, J{\"u}rgen Weese, Josien Pluim, and Marcel Breeuwer.
\newblock On the usability of synthetic data for improving the robustness of deep learning-based segmentation of cardiac magnetic resonance images.
\newblock {\em Medical Image Analysis}, 84:102688, 2023.

\bibitem{man2022review}
Keith Man and Javaan Chahl.
\newblock A review of synthetic image data and its use in computer vision.
\newblock {\em Journal of Imaging}, 8(11):310, 2022.

\bibitem{Jaziri_2024_WACV}
Achref Jaziri, Martin Mundt, Andres Fernandez, and Visvanathan Ramesh.
\newblock Designing a hybrid neural system to learn real-world crack segmentation from fractal-based simulation.
\newblock In {\em Proceedings of the IEEE/CVF Winter Conference on Applications of Computer Vision (WACV)}, pages 8636--8646, January 2024.

\bibitem{xu2023innovative}
Jia Xu, Cheng Yuan, Jiaxuan Gu, Jian Liu, Jiong An, and Qingzhao Kong.
\newblock Innovative synthetic data augmentation for dam crack detection, segmentation, and quantification.
\newblock {\em Structural Health Monitoring}, 22(4):2402--2426, 2023.

\bibitem{SuperviselySyntheticCrackSegmentation}
Supervisely.
\newblock Supervisely synthetic crack segmentation, 2023.

\bibitem{ASADISHAMSABADI2022111590}
Elyas {Asadi Shamsabadi}, Chang Xu, and Daniel~Dias da~Costa.
\newblock Robust crack detection in masonry structures with transformers.
\newblock {\em Measurement}, 200:111590, 2022.

\bibitem{lee2019robust}
Donghan {Lee}, Jeongho {Kim}, and Daewoo {Lee}.
\newblock {Robust Concrete Crack Detection Using Deep Learning-Based Semantic Segmentation}.
\newblock {\em International Journal of Aeronautical and Space Sciences}, 20(1):287--299, March 2019.

\bibitem{RQUOYN_2025}
Johannes Flotzinger, Philipp~J. Rösch, and Thomas Braml.
\newblock {dacl10k}, 2025.

\bibitem{Xu2019}
Hongyan Xu, Xiu Su, Yi~Wang, Huaiyu Cai, Kerang Cui, and Xiaodong Chen.
\newblock Automatic bridge crack detection using a convolutional neural network.
\newblock {\em Applied Sciences}, 9(14), 2019.

\bibitem{Dorafshan2018Dec}
Sattar Dorafshan, Robert~J. Thomas, and Marc Maguire.
\newblock {SDNET2018: An annotated image dataset for non-contact concrete crack detection using deep convolutional neural networks}.
\newblock {\em Data in Brief}, 21:1664--1668, December 2018.

\bibitem{ZOU2012227}
Qin Zou, Yu~Cao, Qingquan Li, Qingzhou Mao, and Song Wang.
\newblock Cracktree: Automatic crack detection from pavement images.
\newblock {\em Pattern Recognition Letters}, 33(3):227--238, 2012.

\bibitem{10.1007/978-981-15-9343-7_36}
Myeongsuk Pak and Sanghoon Kim.
\newblock Crack detection using fully convolutional network in wall-climbing robot.
\newblock In James~J. Park, Simon~James Fong, Yi~Pan, and Yunsick Sung, editors, {\em Advances in Computer Science and Ubiquitous Computing}, pages 267--272, Singapore, 2021. Springer Singapore.

\bibitem{LIU2019139}
Yahui Liu, Jian Yao, Xiaohu Lu, Renping Xie, and Li~Li.
\newblock Deepcrack: A deep hierarchical feature learning architecture for crack segmentation.
\newblock {\em Neurocomputing}, 338:139--153, 2019.

\bibitem{Li_2023}
Yongshang Li, Ronggui Ma, Han Liu, and Gaoli Cheng.
\newblock Real-time high-resolution neural network with semantic guidance for crack segmentation.
\newblock {\em Automation in Construction}, 156:105112, December 2023.

\bibitem{10222276}
Huaqi Tao, Bingxi Liu, Jinqiang Cui, and Hong Zhang.
\newblock A convolutional-transformer network for crack segmentation with boundary awareness.
\newblock In {\em 2023 IEEE International Conference on Image Processing (ICIP)}, pages 86--90, 2023.

\bibitem{CrackwidthDIN}
German~Institute for Standardization Registered Association~({DIN}).
\newblock {Eurocode 2: Design of reinforced and prestressed concrete structures - Part 1-1: General design rules and rules for buildings}.
\newblock Standard, DIN German Institute for Standardization, Berlin, GER, April 2013.

\bibitem{ACICommittee224Cracking}
{ACI Committee 224}.
\newblock {\em Control of cracking in concrete structures}.
\newblock {American Concrete Institute}, Farmington Hills, Mich., 2001.

\bibitem{DIN_1076}
German~Institute for Standardization Registered Association~({DIN}).
\newblock Engineering structures in connection with roads - inspection and test.
\newblock Standard DIN 1076:1999-11, DIN German Institute for Standardization, Berlin, GER, November 1999.

\bibitem{RIEBW}
Federal Highway Research~Institute ({BaSt}).
\newblock {G}uideline for the uniform acquisition, assessment, recording and evaluation of results of structural inspections ({RI-EBW-PRÜF}), February 2017.

\bibitem{ACICommittee201}
{ACI Committee 201}.
\newblock 201.1r-08: Guide for conducting a visual inspection of concrete in service.
\newblock {\em Technical Documents}, 2008.

\bibitem{9D6E4M_2025}
Johannes Flotzinger, Fabian Deuser, and Achref Jaziri.
\newblock {synth-dacl: Synthetic Extensions for dacl10k Dataset}, 2025.

\bibitem{Lin_2017_CVPR}
Tsung-Yi Lin, Piotr Dollar, Ross Girshick, Kaiming He, Bharath Hariharan, and Serge Belongie.
\newblock Feature pyramid networks for object detection.
\newblock In {\em Proceedings of the IEEE Conference on Computer Vision and Pattern Recognition (CVPR)}, July 2017.

\bibitem{maxvit}
Zhengzhong Tu, Hossein Talebi, Han Zhang, Feng Yang, Peyman Milanfar, Alan Bovik, and Yinxiao Li.
\newblock Maxvit: Multi-axis vision transformer.
\newblock In Shai Avidan, Gabriel Brostow, Moustapha Ciss{\'e}, Giovanni~Maria Farinella, and Tal Hassner, editors, {\em Computer Vision -- ECCV 2022}, pages 459--479, Cham, 2022. Springer Nature Switzerland.

\bibitem{Everingham2014ThePV}
Mark Everingham, S.~M.~Ali Eslami, Luc~Van Gool, Christopher K.~I. Williams, John~M. Winn, and Andrew Zisserman.
\newblock The pascal visual object classes challenge: A retrospective.
\newblock {\em International Journal of Computer Vision}, 111:98 -- 136, 2014.

\bibitem{7780719}
Marius Cordts, Mohamed Omran, Sebastian Ramos, Timo Rehfeld, Markus Enzweiler, Rodrigo Benenson, Uwe Franke, Stefan Roth, and Bernt Schiele.
\newblock { The Cityscapes Dataset for Semantic Urban Scene Understanding }.
\newblock In {\em 2016 IEEE Conference on Computer Vision and Pattern Recognition (CVPR)}, pages 3213--3223, Los Alamitos, CA, USA, June 2016. IEEE Computer Society.

\bibitem{8100027}
Bolei Zhou, Hang Zhao, Xavier Puig, Sanja Fidler, Adela Barriuso, and Antonio Torralba.
\newblock Scene parsing through ade20k dataset.
\newblock In {\em 2017 IEEE Conference on Computer Vision and Pattern Recognition (CVPR)}, pages 5122--5130, 2017.

\bibitem{WANG2024110685}
Yuqing Wang, Yun Zhao, and Linda Petzold.
\newblock An empirical study on the robustness of the segment anything model (sam).
\newblock {\em Pattern Recognition}, 155:110685, 2024.

\bibitem{luan2017deepphotostyletransfer}
Fujun Luan, Sylvain Paris, Eli Shechtman, and Kavita Bala.
\newblock Deep photo style transfer, 2017.

\bibitem{hess2021proceduralworldgenerationframework}
Timm Hess, Martin Mundt, Iuliia Pliushch, and Visvanathan Ramesh.
\newblock A procedural world generation framework for systematic evaluation of continual learning, 2021.

\bibitem{park2020contrastivelearningunpairedimagetoimage}
Taesung Park, Alexei~A. Efros, Richard Zhang, and Jun-Yan Zhu.
\newblock Contrastive learning for unpaired image-to-image translation, 2020.

\bibitem{prabhu2023bridgingsim2realgapcare}
Viraj Prabhu, David Acuna, Andrew Liao, Rafid Mahmood, Marc~T. Law, Judy Hoffman, Sanja Fidler, and James Lucas.
\newblock Bridging the sim2real gap with care: Supervised detection adaptation with conditional alignment and reweighting, 2023.

\bibitem{tobin2017domainrandomizationtransferringdeep}
Josh Tobin, Rachel Fong, Alex Ray, Jonas Schneider, Wojciech Zaremba, and Pieter Abbeel.
\newblock Domain randomization for transferring deep neural networks from simulation to the real world, 2017.

\bibitem{kirillov2019panoptic}
Alexander Kirillov, Ross Girshick, Kaiming He, and Piotr Doll{\'a}r.
\newblock Panoptic feature pyramid networks.
\newblock In {\em Proceedings of the IEEE/CVF conference on computer vision and pattern recognition}, pages 6399--6408, 2019.

\bibitem{tu2022maxvit}
Zhengzhong Tu, Hossein Talebi, Han Zhang, Feng Yang, Peyman Milanfar, Alan Bovik, and Yinxiao Li.
\newblock Maxvit: Multi-axis vision transformer.
\newblock In {\em European conference on computer vision}, pages 459--479. Springer, 2022.

\bibitem{Demir2018DeepGlobe2A}
Ilke Demir, Krzysztof Koperski, David Lindenbaum, Guan Pang, Jing Huang, Saikat Basu, Forest Hughes, Devis Tuia, and Ramesh Raskar.
\newblock Deepglobe 2018: A challenge to parse the earth through satellite images.
\newblock {\em 2018 IEEE/CVF Conference on Computer Vision and Pattern Recognition Workshops (CVPRW)}, pages 172--17209, 2018.

\end{thebibliography}

\end{document}